\documentclass[letterpaper, 10 pt, journal, twoside]{IEEEtran}

\usepackage[english]{babel}
\usepackage{microtype}
\usepackage{graphicx}
\usepackage{subfigure}
\usepackage{booktabs}
\usepackage{bm}
\usepackage{subfig}
\usepackage{amsmath}
\usepackage{tensor}
\usepackage{amsfonts}
\usepackage{algorithm}
\usepackage{algorithmic}
\usepackage{url}
\usepackage{amssymb}
\usepackage{amsmath}
\usepackage{breqn}
\usepackage{siunitx}
\usepackage{xcolor}
\usepackage{afterpage}
\usepackage{soul}
\usepackage{tabularx}
\usepackage{subfigure}
\usepackage{multirow}
\usepackage{here} 
\usepackage{relsize}
\usepackage{float}
\usepackage{pifont}
\usepackage{url}
\usepackage{multicol}

\newcommand{\cmark}{\ding{51}}%
\newcommand{\xmark}{\ding{55}}%

\newcommand{\rebuttal}[1]{{\color{black} #1}}

\begin{document}
\title{Physics-Inspired Temporal Learning of Quadrotor Dynamics for Accurate Model Predictive\\Trajectory Tracking}

\author{Alessandro~Saviolo, Guanrui~Li, and Giuseppe~Loianno%
\thanks{This work was supported in part by the NSF CAREER Award under Grant 2145277, in part by DARPA YFA Grant D22AP00156-00, and in part by Technology Innovation Institute, Qualcomm Research, Nokia, and NYU Wireless. Author Giuseppe Loianno
serves as consultant for the Technology Innovation Institute. This arrangement
has been reviewed and approved by the New York University in accordance
with its policy on objectivity in research. (\textit{Corresponding author: Alessandro Saviolo.})}%
\thanks{The authors are with the Tandon School of Engineering, New York University, Brooklyn, NY 11201 USA. e-mail: {\tt\footnotesize \{alessandro.saviolo, lguanrui,loiannog\}@nyu.edu}.}%
\thanks{Digital Object Identifier (DOI): 10.1109/LRA.2022.3192609.}
}

\markboth{IEEE Robotics and Automation Letters. Preprint Version. Accepted July, 2022}{Saviolo \MakeLowercase{\textit{et al.}}: Physics-Inspired Temporal Learning of Quadrotor Dynamics} 

\maketitle

\begin{abstract}
Accurately modeling quadrotor's system dynamics is critical for guaranteeing agile, safe, and stable navigation. The model needs to capture the system behavior in multiple flight regimes and operating conditions, including those producing highly nonlinear effects such as aerodynamic forces and torques, rotor interactions, or possible system configuration modifications. Classical approaches rely on handcrafted models and struggle to generalize and scale to capture these effects. 
In this paper, we present a novel Physics-Inspired Temporal Convolutional Network (\textit{PI-TCN}) approach to learning quadrotor's system dynamics purely from robot experience. Our approach combines the expressive power of sparse temporal convolutions and dense feed-forward connections to make accurate system predictions. In addition, physics constraints are embedded in the training process to facilitate the network's generalization capabilities to data outside the training distribution.
Finally, we design a model predictive control approach that incorporates the learned dynamics for accurate closed-loop trajectory tracking fully exploiting the learned model predictions in a receding horizon fashion.
Experimental results demonstrate that our approach accurately extracts the structure of the quadrotor's dynamics from data, capturing effects that would remain hidden to classical approaches.
To the best of our knowledge, this is the first time physics-inspired deep learning is successfully applied to temporal convolutional networks and to the system identification task, while concurrently enabling predictive control.
\end{abstract}

\begin{IEEEkeywords}
Robot Learning, Model Learning for Control, Optimization and Optimal Control, Aerial Systems.
\end{IEEEkeywords}

\IEEEpeerreviewmaketitle

\vspace{-0.6em}
\section*{Supplementary Material} \label{sec:links}
\noindent \textbf{Video:} \url{https://youtu.be/dsOtKfuRjEk}\\
\noindent \textbf{Code:} \url{https://github.com/arplaboratory/PI-TCN}
\vspace{-0.6em}

\section{Introduction} \label{sec:intro}
\IEEEPARstart{U}{nmanned} Aerial Vehicles (UAVs), such as quadrotors, have become important platforms to help humans solve a wide range of time-sensitive problems including logistics, search and rescue for post-disaster response, and more recently reconnaissance and monitoring during the COVID-19 pandemic. These tasks often require robots to make fast decisions and agile maneuvers in uncertain, cluttered, and dynamic environments.
In these scenarios, to safely control a UAV, it is critical to accurately model the system dynamics to capture the highly nonlinear effects generated by aerodynamic forces and torques, propellers interactions, vibrations, and other phenomena.
However, such effects cannot be easily measured \rebuttal{or modeled}, thus remain hidden. 
Moreover, for some UAV applications, the platform may be extended with external payloads that would significantly change the dynamics by varying the mass and moment of inertia \cite{guanrui2021pcmpc}. 
Overall, failing to model such system changes would result in significant degradation of the flight performance and may cause catastrophic failures.

\begin{figure}[t]
    \centering
    \includegraphics[width=\linewidth, trim=0 675 0 30, clip]{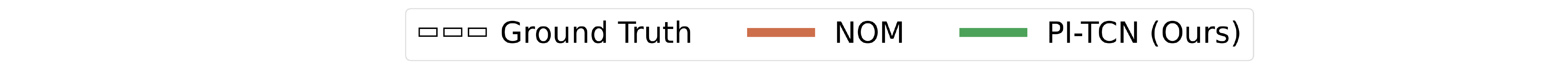}
    \includegraphics[width=0.9\linewidth, trim=-40 100 0 20, clip]{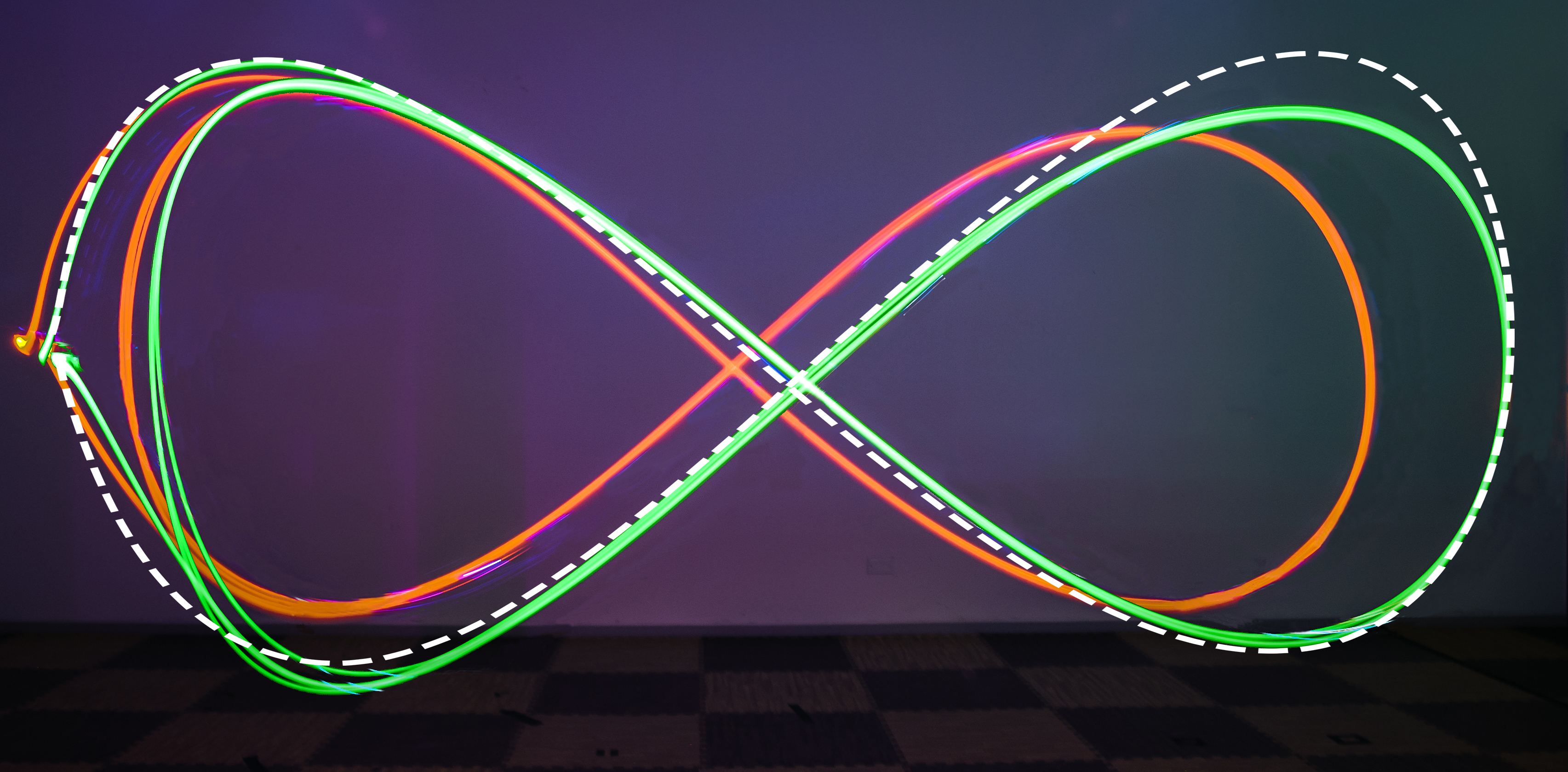}
    \includegraphics[width=\linewidth, trim=-20 0 -20 10, clip]{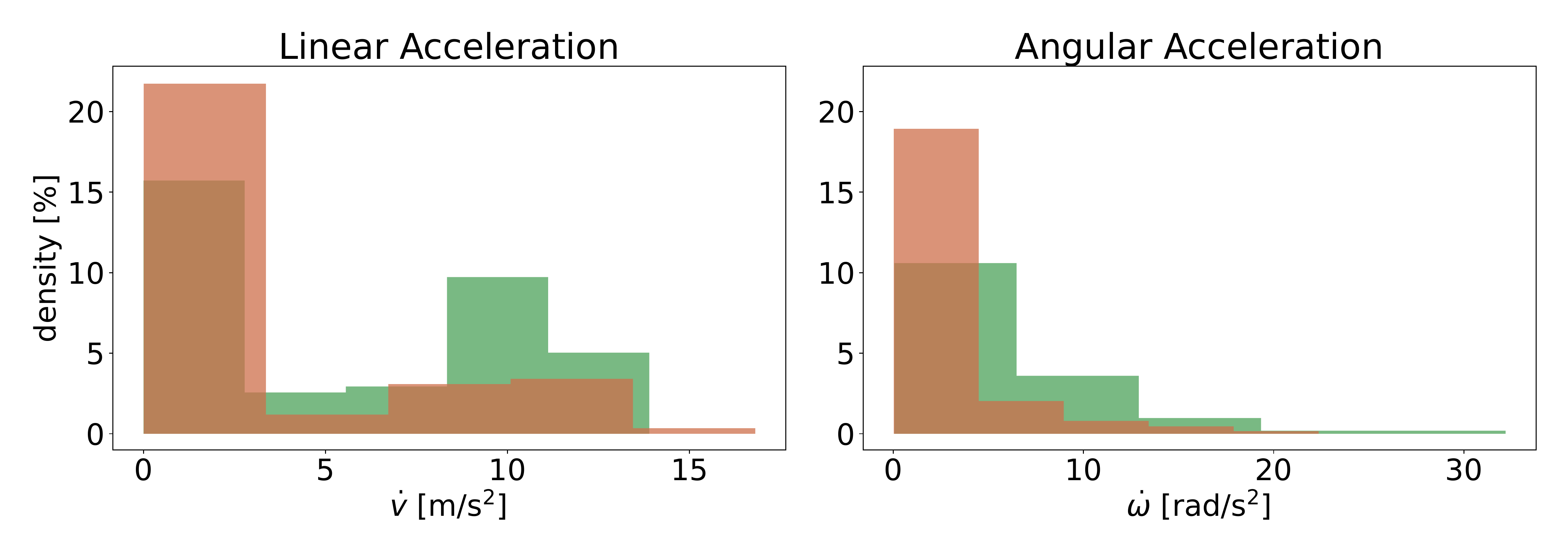}
    \vspace{-2em}
    \caption{\rebuttal{
    \textbf{Top}: Long-exposure photo showing multiple laps over Lemniscate trajectory when using the nominal (NOM) and the proposed (PI-TCN) dynamics. The reference trajectory (white) is approximately projected on the image based on real-world experiments for illustrative purposes. \textbf{Bottom}: Density histograms of accelerations. The quadrotor reaches accelerations up to $13~\SI{}{\meter\per\second\squared}$ and $32~\SI{}{\radian\per\second\squared}$.} \label{fig:tracking_error}}
    \vspace{-20pt}
\end{figure}

Classic modeling of the quadrotor's dynamics is performed using physics-based principles approaches which result in nonlinear \rebuttal{ordinary} differential equations \cite{lee2010empiricalmethod, bouabdallah2007bem, sanchez2018kinconstr, duong2021tuninghamilton, das2009lagmechanics}.
However, these nominal models only approximate the actual system dynamics and do not take into account external effects caused by aggressive maneuvers or system modifications.
To circumvent this issue, recent works have investigated data-driven approaches for modeling system dynamics. 
Several methodologies exist, from combining nominal models with learned residual terms \cite{shi2019neurallander,torrente2021datadrivenmpc,bauersfeld2021neurobem} to fully predicting the system dynamics using neural networks \cite{punjani2015helicopter,bansal2016fulldynnnlearning,kaufmann2020deepacrob,looper2021tcnbadassumptions}.

Successfully learning accurate dynamics directly from data would have a terrific impact on the development of new robotic systems, enabling fast modeling and high-performance control in multiple operating conditions with the potential to scale to any type of platform. 
Therefore, the goal of this work is to fully leverage the expressive power of deep neural networks to extract the quadrotor's system dynamics purely from data.
We propose a Physics-Inspired Temporal Convolutional Network (PI-TCN) that combines a temporal convolutional network that encodes time-correlated features from a history of past states and control inputs with a multi-layer perceptron. The latter decodes the latent temporal representation into an accurate prediction over the quadrotor's dynamics. The learned dynamical model can then be employed for accurate predictive trajectory tracking, as shown in Figure~\ref{fig:tracking_error}. 
Following the recent exciting developments in physics-inspired learning \cite{raissi2019pinn, moseley2020solvingwaveeq, yang2021pi_rnn}, we constrain the network's predictions to be consistent with physical laws by introducing a physics-inspired loss term during training. The composite loss fosters the network's generalization performance on data outside the training distribution, leading to a more stable convergence.

This paper presents multiple contributions. First, we design and present PI-TCN for learning quadrotor dynamics. To date, this is the first time physics-inspired learning is applied to temporal convolutional networks since most of the approaches have only focused on feed-forward and recurrent neural networks \cite{moseley2020solvingwaveeq, yang2021pi_rnn}.
Second, we perform an extensive evaluation by comparing the predictive performance of the proposed methodology against nominal, \rebuttal{residual,} and learning methods on unseen test maneuvers in both simulation and real-world settings. \rebuttal{We experimentally validate the network's design choices and propose several ablation studies to evaluate their roles in capturing quadrotor's dynamics} from data, even highly nonlinear effects that remain hidden to classical \rebuttal{and residual} approaches.
Finally, we propose a model predictive control framework that leverages the learned dynamical model and we extensively test it to track multiple aggressive maneuvers.

\section{Related Works}
Consider a dynamical system with state $\mathbf{x}$ and control input $\mathbf{u}$. Solving the system identification problem requires to find a function $h$, parameterized by $\boldsymbol{\theta}$, that maps from state-control space to state-derivative space:
\begin{equation*} \label{eq:system_identification}
\dot{\mathbf{x}} = 
h(\mathbf{x}, \mathbf{u}; \boldsymbol{\theta})
.
\end{equation*}
Based on $h$ formulation, system identification methods can be categorized into nominal, residual, and learning methods.

\textbf{Nominal Methods},
Nominal methods formulate $h$ using \rebuttal{physics-based principles} \cite{lee2010empiricalmethod}. Such models have been further refined by using blade element momentum theory \cite{bouabdallah2007bem}, kinematic constraints \cite{sanchez2018kinconstr}, Hamiltonian and Lagrangian mechanics \cite{duong2021tuninghamilton, das2009lagmechanics}.
In general, these nominal models are computationally efficient and describe the quadrotor's system dynamics well in low-speed regimes and basic platforms, where external forces and torques are negligible. However, as speed increases or \rebuttal{additional} payloads are applied to the platform, external complex effects increase as well, significantly degrading the flying performance \cite{saviolo2021autotune, alkayas2022extdisturbances}. Moreover, nominal models %
\rebuttal{depend on the physical system parameters}. Parameter identification approaches can be used to empirically identify their values \cite{wueest2018estimation, svacha2020inertiaest}. However, uncertainties remain due to the nonlinearity of external effects that make them difficult to be estimated.

\textbf{Residual Methods}.
Inspired by the success of deep neural networks, several works proposed to combine nominal methods with data-driven techniques for learning the residual terms not being modeled by physics-based principles.
\cite{shi2019neurallander} extended the nominal model with a residual term predicted by a feed-forward neural network to capture the aerodynamic forces affecting the linear acceleration of a quadrotor during landing. However, the generalization capabilities of the proposed residual method remain unknown due to the limited number of experiments involving a single trajectory used both for training and testing.
\cite{torrente2021datadrivenmpc} combined the nominal model with residual forces predicted by a Gaussian process. A major drawback of Gaussian process regression is computational complexity. Since these models are non-parametric, their complexity increases with the size of the training set. This implies the need to carefully choose a subset of the training data which best represents the true dynamics. However, since the dynamics are unknown, selecting these points might be challenging.
\cite{bauersfeld2021neurobem} refined the nominal model leveraging blade element momentum theory to better characterize motor forces when affected by aerodynamic effects. Subsequently, they combined the model with learned residual force and torque terms represented by deep neural networks. The proposed residual model benefits from the generalization guaranteed by physics-based principles and the flexibility of learning-based function approximations. However, as for all residual methods, the relationship between the complex effects and the true dynamics is assumed to be well known and the learning techniques are adopted only to predict the residual additive terms. 
\rebuttal{Conversely, we relax the hard physical constraint imposed by embedding the nominal model and train the network to extract the true dynamics from the data using physical laws only as a soft constraint to direct the training convergence to well-generalizable solutions.
Therefore, the network is concurrently able to exploit the physical model and explore the loss landscape.
}

\textbf{Learning Methods}.
Inspired by the promising results of residual methods and the rather simple data collection procedure that does not involve any expensive manual labeling, several works have been investigated to entirely learn the governing equations of system dynamics from data.
For example, \cite{bansal2016fulldynnnlearning} adopted a shallow feed-forward neural network for learning the full system dynamics for quadrotor flight. While feedforward neural networks allow modeling highly complex phenomena, they are not designed for learning time-correlated features.
When modeling time-series data, recurrent neural networks provide better architectures.
However, recurrent neural networks have some important limitations, from vanishing and exploding gradient problems to the difficulty to process long sequences. Such limitations make these networks complex to properly train and poorly suited for online robotics applications \cite{pascanu2013rnndifficult}. 
Recently, convolution-based approaches have emerged as a superior alternative to Recurrent Neural Networks (RNNs) \cite{bai2018tcnbeatrnn}. \cite{lea2017tcn} introduced the temporal convolutional network (TCN) to perform fine-grained action segmentation. Unlike RNNs, TCNs take advantage of asynchronous and parallel convolution operations, avoid gradient instability problems, and offer flexible receptive field size thus better control of the model's memory size while still inherently accounting for temporal data structures like RNNs. Thanks to their favorable characteristics, TCNs have then been successfully employed in multiple sequences and time-series modeling tasks \cite{kaufmann2020deepacrob, looper2021tcnbadassumptions, borovykh2018conditional, luo2019tcnapplication}. 
Related to quadrotor control, \cite{kaufmann2020deepacrob} trained multiple TCNs to learn directly from raw sensory data an end-to-end policy for performing acrobatic maneuvers. 
For quadrotor's system identification, \cite{looper2021tcnbadassumptions} incorporated the entire system in a TCN and demonstrated the utility and applicability of these network's architectures for learning the full system dynamics. However, the learned model is obtained from data in a limited flight regime \rebuttal{and the generalization beyond the training distribution is not considered.} It also strongly relies on accessing future control inputs which are certainly useful, but impractical in real-world scenarios. In addition, the learned model is not used within a receding horizon control framework to exploit its predictive nature.  Conversely, \rebuttal{in this work, we} do not make any assumptions over future control \rebuttal{and demonstrate the learned dynamics for accurate closed-loop trajectory tracking in real-world experiments}.

Existing learning methods either decouple linear and angular accelerations or only estimate the former. This limits the predictive performance of the neural networks that can capture the hidden dependencies that bound forces and torques. Instead, we predict the entire dynamics jointly.

\section{Methodology}
We first introduce the nominal model of the quadrotor's system dynamics that will be used by the proposed physics-inspired loss function and then formulate the system identification problem using PI-TCN. 
Table~\ref{tab:symbols} lists the relevant variables used in the paper. %Figure~\ref{fig:system_convention} illustrates the physical configuration of the quadrotor.

\subsection{Nominal Formulation}
Nominal methods model the quadrotor's system dynamics by using nonlinear \rebuttal{ordinary} differential equations.
Specifically, consider the quadrotor system modeled by the state vector $\arraycolsep=2pt \mathbf{x}=\begin{bmatrix}\mathbf{p}^\top&\mathbf{v}^\top&\mathbf{q}^\top&\boldsymbol{\omega}^\top\end{bmatrix}^\top$ and the control input $\mathbf{u}$, then the quadrotor's nominal dynamics evolve as follows
\begin{equation} \label{eq:empirical_method}
    \dot{\mathbf{x}} =
    \begin{bmatrix}
    \dot{\mathbf{p}} \\[1ex]
    \dot{\mathbf{v}} \\[1ex]
    \dot{\mathbf{q}} \\[1ex]
    \dot{\boldsymbol{\omega}}
    \end{bmatrix} =
    \begin{bmatrix}
    \mathbf{v} \\[1ex]
    \frac{1}{m} ( \mathbf{q} \odot f ) + \boldsymbol{g} \\[1ex]
    \frac{1}{2} ( \mathbf{q} \odot \boldsymbol{\omega} ) \\[1ex]
    \mathbf{J}^{-1} (\boldsymbol{\tau} - \boldsymbol{\omega} \times \mathbf{J} \boldsymbol{\omega})
\end{bmatrix} =
    h_{\text{Nom}}(\mathbf{x}, \mathbf{u})
    ,
\end{equation}
\rebuttal{where $\mathbf{J} = diag(J_{xx}, J_{yy}, J_{zz})$ is the diagonal moment of inertia matrix, $\mathbf{g} = \arraycolsep=2pt \begin{bmatrix} 0 & 0 & -g \end{bmatrix}^\top$ is the gravity vector, and the collective thrust $f$ and torque $\tau$ of the quadrotor are defined as
\begin{equation} \label{eq:torque}
    f = k_{f} \sum_{i=0}^3 u_i^2,\hspace{1em}
    \boldsymbol{\tau} = 
    \begin{bmatrix}
    k_{f} l (u_0^2 + u_1^2 - u_2^2 - u_3^2) \\[1ex]
    k_{f} l (-u_0^2 + u_1^2 + u_2^2 - u_3^2) \\[1ex]
    k_{\tau} (u_0^2 - u_1^2 + u_2^2 - u_3^2)
    \end{bmatrix}
    .
\end{equation}
The parameters $J_{xx}, J_{yy}, J_{zz}, m, k_{f}, k_{\tau}, l$ are related to the physical system and strictly define the nominal model $h_{\text{Nom}}$. 
Accurately identifying their values is key for guaranteeing high-performance flight control while using nominal dynamics. 
However, precisely modeling the system's parameters is very difficult due to the nonlinearity of external effects that make the estimation process difficult.
}

\subsection{Model Learning}
In this work, we approximate $h$ using a physics-inspired temporal convolutional network and leverage past flight states and control inputs to predict the quadrotor's full dynamic state.
Formally, the state-derivative at time $i$ is given by
\begin{equation} \label{eq:quadrotor_system}
    \dot{\mathbf{x}}_i =
    h_{\text{PI-TCN}}(\mathbf{X}_i, \mathbf{U}_i; \boldsymbol{\theta}),
\end{equation}
where $\arraycolsep=1.5pt \mathbf{X}_i=\begin{bmatrix}\mathbf{x}_{i-T}^\top&\hdots&\mathbf{x}_{i}^\top\end{bmatrix}^\top$ and $\arraycolsep=1.5pt \mathbf{U}_i=\begin{bmatrix}\mathbf{u}_{i-T}^\top&\hdots&\mathbf{u}_{i}^\top\end{bmatrix}^\top$ are histories of states and control inputs of length $T$, while $\boldsymbol{\theta}$ represents the network's parameters.
%
% and $\mathbf{U}_i = \begin{bmatrix}\mathbf{u}_{i-T} \\ \mathbf{u}_{i-T+1} \\ \vdots \\ \mathbf{u}_{i} \end{bmatrix}$ are the augmented state and input vectors with the state history from time $i-T$ to time $i$.
%
% \begin{equation*}
%     \mathbf{X}_i = 
%     \begin{bmatrix}
%         \mathbf{x}_{i-T} \\
%         \mathbf{x}_{i-T+1} \\
%         \vdots \\
%         \mathbf{x}_{i}
%     \end{bmatrix},~
%     \mathbf{U}_i = 
%     \begin{bmatrix}\mathbf{u}_{i-T} \\
%         \mathbf{u}_{i-T+1} \\
%         \vdots \\
%         \mathbf{u}_{i}
%     \end{bmatrix}
%     .
% \end{equation*}
%
Therefore, solving the system identification problem corresponds to learning the parameters $\boldsymbol{\theta}$ of the network $h_{\text{PI-TCN}}$.

\begin{table}[t]
    \caption{\label{tab:symbols}}
    \vspace{-0.75em}
    \caption*{\scshape Notation Table}
    \vspace{-0.5em}
    \centering
    \begin{tabular}{l l}
        \toprule
        \midrule
        $\mathcal{I}, \mathcal{B}$ & inertial, body frame \\
        $m$ & mass of quadrotor in $\mathcal{I}$ \\
        $\mathbf{p}\in\mathbb{R}^3$ & position of quadrotor in $\mathcal{I}$ \\
        $\mathbf{v}\in\mathbb{R}^3$ & linear velocity of quadrotor in $\mathcal{I}$ \\
        $\mathbf{q}\in\mathbb{R}^4$ & orientation of quadrotor with respect to $\mathcal{I}$ \\
        $\boldsymbol{\omega}\in\mathbb{R}^3$ & angular velocity of quadrotor in $\mathcal{B}$ \\
        $\mathbf{u}\in\mathbb{R}^4$ & motor commands generated by quadrotor's controller \\
        $\dot{\mathbf{v}}\in\mathbb{R}^3$ & linear acceleration of quadrotor in $\mathcal{B}$ \\
        $\dot{\boldsymbol{\omega}}\in\mathbb{R}^3$ & angular acceleration of quadrotor in $\mathcal{B}$ \\
        $f\in\mathbb{R}$ & total thrust of quadrotor \\
        $\boldsymbol{\tau}\in\mathbb{R}^3$ & torque of quadrotor in $\mathcal{B}$ \\
        $\mathbf{J}\in\mathbb{R}^{3\times3}$ & diagonal moment of inertia matrix of quadrotor \\
        $k_{f}$ & rotor thrust constant \\
        $k_{\tau}$ & rotor torque constant \\
        $l$ & length of the quadrotor arm \\
        $g$ & gravity constant\\
        $\odot$ & quaternion-vector product \\
        \midrule
        \bottomrule
    \end{tabular}
    \vspace{-0.5em}
\end{table}

\begin{figure*}
    \centering
    \includegraphics[width=\textwidth, trim=30 110 30 150, clip] {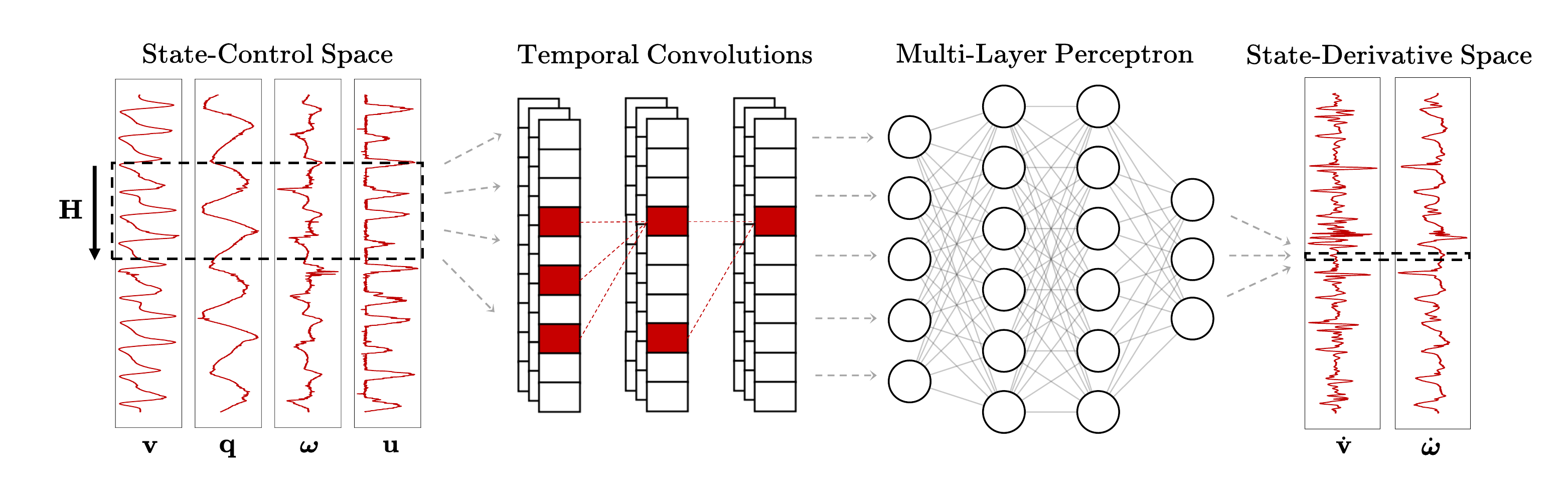}
    \vspace{-2em}
    \caption{PI-TCN's architecture. The network receives an history of past flight inputs and states from time $i-T$ to time $i$. A TCN extracts a sequence of time-correlated features, which are then processed by a MLP to predict the full system dynamics.
    \label{fig:network_architecture}}
    \vspace{-1.3em}
\end{figure*}

The proposed network $h_{\text{PI-TCN}}$, illustrated in Figure~\ref{fig:network_architecture}, consists of two sub-networks, a temporal convolutional network (TCN) and a multi-layer perceptron (MLP). 
The TCN extracts time-correlated features from a sequence of past flight states and control inputs and outputs a compact hidden-state vector. Such hidden state is passed in input to the MLP which predicts the quadrotor's full system dynamics.
Such encoder-decoder architecture fully leverages the qualities of the TCN and MLP models. TCNs are sparse networks defined by dilated (causal) convolutional layers, which allow to process long history sequences in parallel while encoding the temporal structure of the input time series in their output feature vector. Conversely, MLPs are dense networks, which makes them better suited for making predictions from a compact hidden state representation.

We provide as input to the network a history of states and control inputs of length $T$, with samples temporarily equally spaced (see supplementary material for an in-depth study of the network's performance with different history lengths).
Each state consists of linear velocity $\mathbf{v}$, attitudes $\mathbf{q}$, angular velocity $\boldsymbol{\omega}$, and the control inputs are the motor commands $\mathbf{u}$. Therefore, the network's input is a tensor of shape $14 \times T$. 
The network's output is the full dynamic state of the quadrotor. However, since position and orientation derivatives are already provided as input, we restrict the prediction to linear and angular accelerations. 
Consequently, the network's output is a $6 \times 1$ tensor. Note that we do not provide position information as the linear and angular accelerations are position-independent.

\subsection{Physics-Inspired Loss}
Learning the dynamics purely from data poses the challenge to make the network generalizable outside the training distribution. However, it is necessary to concurrently guarantee that the network matches physical principles.
Motivated by this observation and inspired by~\cite{raissi2019pinn, moseley2020solvingwaveeq, yang2021pi_rnn}, where it is clearly shown the benefits of constraining the network to physical principles, we embed physics constraints in the training process by including the physics laws in Eq.~(\ref{eq:empirical_method}) in the loss function. Specifically, at each training iteration, we minimize the composite loss
\begin{equation} \label{eq:composite_loss}
\mathcal{L} = \mathcal{L}_{\text{MSE}} + \rebuttal{\lambda} \mathcal{L}_{\text{PI}},
\end{equation}
where $\mathcal{L_{\text{MSE}}}$ is the mean squared error prediction loss between the training labels and the network's predictions, $\mathcal{L_{\text{PI}}}$ is the physics-inspired loss between the physics laws' solution and the network's predictions, \rebuttal{and $\lambda$ is a hyper-parameter that should reflect how confident we are in the physical constraints of our system. If we have access to an unreliable physical model, $\lambda$ should be small to let the neural network fully explore the loss landscape. On the other hand, if we are confident in the available physical model, $\lambda$ can be large to fully exploit the prior.}
The loss functions are
\begin{equation} \label{eq:mse_loss}
\mathcal{L}_{\text{MSE}} = \frac{1}{|B_T|} \sum _{i=1} ^{|B_T|} || \bar{h}(\mathbf{x}_i, \mathbf{u}_i) - h_{\text{PI-TCN}}(\mathbf{X}_i, \mathbf{U}_i; \boldsymbol{\theta}) ||\\
,
\end{equation} \label{eq:pi_loss}
\begin{equation}
\mathcal{L}_{\text{PI}} = \frac{1}{|B_P|} \sum _{j=1} ^{|B_P|}  || h_{\text{Nom}}(\mathbf{x}_j, \mathbf{u}_j) - h_{\text{PI-TCN}}(\mathbf{X}_j, \mathbf{U}_j; \boldsymbol{\theta}) ||
,
\end{equation}
where $\bar{h}$ \rebuttal{gives} the label for the data point $\left(\mathbf{x}_i, \mathbf{u}_i\right)$, $B_T$ is a batch of training data points, $B_P$ is a batch of points sampled from the entire input space.
While $\mathcal{L_{\text{MSE}}}$ ensures that the network learns the full dynamics purely from data, $\mathcal{L_{\text{PI}}}$ constraints the predictions to match the underlying equations derived from physics-based principles. $\mathcal{L_{\text{PI}}}$ gives the network a physical interpretation of its internal states and can be viewed as an unsupervised regularizer that fosters the network's generalization performance by stabilizing the training process.

We further improve the training process convergence by adopting a curriculum learning strategy. We train the network for half the training process by \rebuttal{setting $\lambda = 0$}.
%only using $\mathcal{L_{\text{MSE}}}$. 
This ensures that the network fully explores the optimization space in a self-supervised fashion. Then, we restart the training using \rebuttal{$\lambda = 1$} for the remaining training iterations to stabilize the training process convergence.
At every training iteration, the physics-inspired loss is computed over a batch of points sampled from the entire state-input space. Selecting these points is trivial if the considered past flight history is unitary. However, in our scenario, we would need to randomly generate consistent sequences of points from the state-input space. Therefore, in this work, we extract a batch of $|B_P|$ points from the state-input space before starting the training process and use it at every training iteration.

\subsection{Control Design} \label{sec:mpc}
We consider the quadrotor trajectory tracking problem, where the platform is required to follow a given desired trajectory of states $\mathbf{x}_{des, i}$ and inputs $\mathbf{u}_{des, i}$. We combine the predictive nature of Model Predictive Control (MPC) and the proposed network to accurately track trajectories while respecting physical or dynamic constraints. MPC formulates an optimization problem that finds a sequence of inputs within a fixed time horizon with $N$ discretized steps by optimizing a given objective function. %The optimization is also subject to the system dynamics constraints. 
The optimization problem is formulated to minimize $\Tilde{\mathbf{x}}_i=\mathbf{x}_{des,i} -\mathbf{x}_i$ and $\Tilde{\mathbf{u}}_i=\mathbf{u}_{des,i} -\mathbf{u}_i$, which are the errors between the desired state and input and the actual state and input. 
Formally, the MPC framework is defined as follows
\begin{equation}
    \begin{split}
        \min_{\mathbf{u}_{0},\cdots,\mathbf{u}_{N-1}}&
        \frac{1}{2}\Tilde{\mathbf{x}}_N^\top \mathbf{Q}_x\Tilde{\mathbf{x}}_N + \sum_{i=0}^{N-1}
        \left(
        \frac{1}{2}\Tilde{\mathbf{x}}_i^\top \mathbf{Q}_x\Tilde{\mathbf{x}}_i + \frac{1}{2}\Tilde{\mathbf{u}}_i^\top \mathbf{Q}_u\Tilde{\mathbf{u}}_i
        \right)\\
        %\min_{\mathbf{u}_{0},\mathbf{u}_{1},\cdots,\mathbf{u}_{N-1}}& ~\sum_{k=0}^{N-1}  L\left(\mathbf{x}_{k},\mathbf{u}_{k}\right) + Q\left(\mathbf{x}_N\right)\\
        \text{s.t.}&~\mathbf{X}_{i+1} = \hat{h}_{\text{PI-TCN}}(\mathbf{X}_{i},\mathbf{U}_{i}; \boldsymbol{\theta}) ,~\forall~i = 0,\cdots,N-1\\
&~g(\mathbf{x}_{i},\mathbf{u}_{i})\leq 0,
\end{split}\label{eq:standard_MPC}
\end{equation}
where $\mathbf{Q}_x$, $\mathbf{Q}_u$ are constant positive diagonal weight matrices in the cost function and $\hat{h}_{\text{PI-TCN}}(\mathbf{X}_{i},\mathbf{U}_{i}; \boldsymbol{\theta})$ is the system dynamics constraint defined as
\begin{equation} \label{eq:mpc_system_dyn}
    \hat{h}_{\text{PI-TCN}}(\mathbf{X}_{i},\mathbf{U}_{i}; \boldsymbol{\theta}) = 
    RK(h_{\text{PI-TCN}}(\mathbf{X}_i,\mathbf{U}_i; \boldsymbol{\theta}))
    .
\end{equation}
%
% \begin{equation} \label{eq:mpc_system_dyn}
%     \hat{h}_{\text{PI-TCN}}(\mathbf{X}_{i},\mathbf{U}_{i}) = 
%     \begin{bmatrix}
%     \mathbf{x}_{i-T+1} \\
%     \vdots \\
%     \mathbf{x}_{i} \\
%     RK(h_{\text{PI-TCN}}(\mathbf{X}_i,\mathbf{U}_i))
%     \end{bmatrix}.
% \end{equation}
%
The $RK$ in Eq.~(\ref{eq:mpc_system_dyn}) is the Runge-Kutta 4th order numerical integration function that numerically integrates the states derivative, given by the PI-TCN model, within a given time step. The optimization occurs with initial condition $\mathbf{x}_0$ while respecting system dynamics $\mathbf{X}_{i+1}=\hat{h}_{\text{PI-TCN}}(\mathbf{X}_{i},\mathbf{U}_{i}; \boldsymbol{\theta})$ and additional state and input constraints $g\left(\mathbf{x}_i,\mathbf{u}_i\right)\leq0$ such as actuator or perception constraints.

\section{Experimental Setup} \label{sec:exp_setup}

\subsection{System}
We learn the dynamical system of a $250~\si{g}$ quadrotor equipped with a Qualcomm\textsuperscript{\textregistered} Snapdragon\textsuperscript{\texttrademark} board and four brushless motors based on our previous work~\cite{loianno2017aggressmaneuver}. We run the MPC on a laptop computer \rebuttal{at $100~\si{Hz}$} and send \rebuttal{via Wi-Fi} the desired body rates and collective thrust to the low-level quadrotor control.
We develop and train the neural networks using PyTorch and implement the controller using CasADi \cite{casadi} and ACADOS \cite{acados}. However, since CasADi builds a static computational graph and postpones the processing of the data, it is not compatible with PyTorch which directly performs the computations using the data. We solve this issue by implementing our network directly in CasADi.

\subsection{Collected Data}
We collect the data by controlling the quadrotor in a series of flights both in simulation and in the real world, resulting in two datasets with analogous trajectories. The simulated flights are performed in the Gazebo simulator, while the real-world flights are performed in an indoor environment $10\times6\times4~\si{m^3}$ at the Agile Robotics and Perception Lab (ARPL) at the New York University. The environment is equipped with a Vicon
%\footnote{https://www.vicon.com/} 
motion capture system that allows recording accurate position and attitude measurements at $100~\si{Hz}$. Additionally, we record the onboard motor speeds.
Each dataset consists of $68$ trajectories with a total of $58^\prime~03^{\prime\prime}$ flight time. 
The trajectories range from straight-line accelerations to circular motions, but also parabolic maneuvers and lemniscate trajectories. All the trajectories are performed for any axis combination (i.e., $x-y$, $x-z$, $y-z$) and with different speeds and accelerations.
To capture the complex effects induced by aggressive flight, we push the quadrotor to its physical limits reaching speeds of $6~\si{\meter\per\second}$, linear accelerations of $18~\si{\meter\per\second\squared}$, \rebuttal{angular accelerations of $54~\si{\radian\per\second\squared}$,} and motor speeds of $16628~\si{rpm}$.
We recover unobserved accelerations from velocity measurements filtered by a UKF. Moreover, we filter the recorded attitudes and motor speeds measurements using a $4\textsuperscript{th}$ order Butterworth lowpass filter with a cutoff frequency of $5$.
We scale the motor speed data by multiplying them by $0.001$ such that the scale of all the data components is equally distributed (i.e., the neural network will give equal importance to all data components).
Finally, we randomly select $60$ trajectories for training, while using the remaining $8$ for testing (Figure~\ref{fig:test_trajs}). 
\rebuttal{See supplementary material for more details on the collected data.}
%The network is trained over the shuffled training set.

\subsection{Baselines}
We implement PI-TCN's architecture as a TCN with $4$ hidden layers each of size $16$ and an MLP with $3$ hidden layers of sizes $64$, $32$, $32$. Each layer of the TCN is stacked with a ReLU activation function, batch normalization, and a Dropout regularizer with rate $10 \%$, whereas each layer of the MLP is stacked with a ReLU activation function. We provide as input to the network a history of states and control inputs of length $T=20$, with samples temporary equally spaced with $\delta t = 10~\SI{}{\milli\second}$, resulting in a temporal window of the system evolution over the past $200~\SI{}{\milli\second}$. We train PI-TCN on the real-world dataset for $10000$ epochs using Adam stochastic gradient descent algorithm, batches of $|B_T| = |B_P| = 1024$ samples, and a constant learning rate of $10^{-4}$.

\rebuttal{We validate PI-TCN by comparing its predictive performance on unseen real-world trajectories against the nominal model in Eq.~(\ref{eq:quadrotor_system}) (\textit{NOM}), a residual TCN (\textit{RES-TCN}), a TCN trained without the physics-inspired loss (\textit{MSE-TCN}), and a multi-layer perceptron (\textit{PI-MLP}). We keep the architecture for all TCN and MLP models the same as PI-TCN for a fair comparison. 
RES-TCN and MSE-TCN are trained on the real-world dataset using the MSE loss in Eq.~(\ref{eq:mse_loss}) for $10000$ epochs, Adam stochastic gradient descent algorithm, batch sizes $|B_T| = |B_P| = 1024$, and a constant learning rate of $10^{-4}$. The same training process is used to train PI-MLP but using the composite loss in Eq.~(\ref{eq:composite_loss}). Moreover, we train PI-TCN and PI-MLP only on simulation data (\textit{PI-TCN*}, \textit{PI-MLP*}) to study the generalization capabilities of the learned models to significant domain changes (\textit{sim-to-real} in this case).
See supplementary material for more details on the baselines.}

% We validate PI-TCN by comparing its predictive performance over unseen data against the nominal model in Eq.~(\ref{eq:quadrotor_system}) and a multi-layer perceptron (PI-MLP). 
% %
% The nominal model is defined by a set of hyper-parameters related to the physical platform, namely the thrust and torque coefficients, the inertia matrix, and the center of mass. We estimate the thrust and torque coefficients by using an RC-benchmark Thrust Stand and we approximate the inertia matrix and the center of mass through precisely generated CAD models as in \cite{wueest2018estimation}. 
% %
% The \rebuttal{PI-MLP} consists of 3 hidden layers of sizes $64$, $32$, $32$, each stacked with a ReLU activation function (equivalently to the decoder of PI-TCN). The network is trained using the composite loss in Eq.~(\ref{eq:composite_loss}) for $10000$ epochs, Adam stochastic gradient descent algorithm, batch sizes $|B_T| = |B_P| = 1024$, and a constant learning rate of $10^{-4}$.

\section{Results} \label{sec:exp_results}
We design our evaluation procedure to address the following questions. 
i) Can model learning approaches extract quadrotor's system dynamics from \rebuttal{robot experience}?
ii) How does PI-TCN compare to the baseline models on simulated and real-world data? 
iii) What is the contribution of \rebuttal{the learned dynamics} in a closed-loop tracking task?

\begin{table*}[t]
    \centering
    \caption{\label{tab:prediction_error}}
    \vspace{-0.75em}
    \caption*{\scshape Comparison of PI-TCN with All the Baselines}
    \rebuttal{
    \vspace{-0.5em}
    \begin{tabular}{c c c c c c c c c c c c c}
    \toprule\toprule
    & $\dot{\mathbf{v}}_{\max}~[\SI{}{\meter\per\second\squared}]$ &
    $\dot{\boldsymbol{\omega}}_{\max}~[\SI{}{\radian\per\second\squared}]$ &
    NOM & RES-TCN & MSE-TCN & PI-MLP* & PI-TCN* & PI-MLP & PI-TCN\\
    \midrule
    Ellipse\_1 & $ 1.10 $ & $ 5.26 $ & $ 0.40 $ & $ 0.07 $ & $ 0.10 $ & $ 0.48 $ & $ 0.36 $ & $ 0.25 $ & $ \textbf{0.07} $ \\
    Ellipse\_2 & $ 5.89 $ & $ 7.34 $ & $ 1.49 $ & $ 0.21 $ & $ 0.22 $ & $ 2.61 $ & $ 0.92 $ & $ 0.73 $ & $ \textbf{0.16} $ \\
    WarpedEllipse\_1 & $ 4.92 $ & $ 4.76 $ & $ 0.69 $ & $ 0.12 $ & $ 0.21 $ & $ 0.93 $ & $ 0.53 $ & $ 0.43 $ & $ \textbf{0.11} $ \\
    WarpedEllipse\_2 & $ 9.99 $ & $ 12.76 $ & $ 2.02 $ & $ 0.99 $ & $ 0.48 $ & $ 2.14 $ & $ 1.17 $ & $ 1.01 $ & $ \textbf{0.20} $ \\
    Lemniscate & $ 13.14 $ & $ 31.98 $ & $ 9.16 $ & $ 1.88 $ & $ 1.39 $ & $ 9.19 $ & $ 5.73 $ & $ 1.62 $ & $ \textbf{0.39} $ \\
    ExtendedLemniscate & $ 7.76 $ & $ 27.95 $ & $ 1.34 $ & $ 1.01 $ & $ 0.69 $ & $ 2.74 $ & $ 0.90 $ & $ 0.49 $ & $ \textbf{0.19} $ \\
    Parabola & $ 5.91 $ & $ 6.57 $ & $ 1.03 $ & $ 0.88 $ & $ 0.33 $ & $ 2.17 $ & $ 0.77 $ & $ 0.38 $ & $ \textbf{0.15} $ \\
    TransposedParabola & $ 17.86 $ & $ 54.90 $ & $ 7.94 $ & $ 1.48 $ & $ 1.01 $ & $ 9.01 $ & $ 3.46 $ & $ 1.59 $ & $ \textbf{0.51} $ \\
    \midrule
    Inference $[\SI{}{\milli\second}]$ & & & $\textbf{0.001}$ & $1.735$ & $1.735$ & $0.229$ & $1.735$ & $0.229$ & $1.735$\\
    \bottomrule\bottomrule
    \end{tabular}
    }
    \vspace{-1em}
\end{table*}

% \begin{figure*}[t]
% \centering
% \subfigure{\includegraphics[width=0.15\textwidth, trim=380 20 300 100,
% clip]{images/trajectories/Ellipse.pdf}}
% \subfigure{\includegraphics[width=0.15\textwidth, trim=380 20 300 100,
% clip]{images/trajectories/WarpedEllipse.pdf}}
% \subfigure{\includegraphics[width=0.15\textwidth, trim=380 20 300 100, clip]{images/trajectories/Parabola.pdf}}
% \subfigure{\includegraphics[width=0.15\textwidth, trim=380 20 300 100, clip]{images/trajectories/Lemniscate.pdf}}
% \subfigure{\includegraphics[width=0.15\textwidth, trim=380 20 300 100, clip]{images/trajectories/ExtendedLemniscate.pdf}}
% \subfigure{\includegraphics[width=0.15\textwidth, trim=380 20 300 100, clip]{images/trajectories/TransposedParabola.pdf}}
% \vspace{-0.75em}
% \caption{Testing trajectories considered in this work. \label{fig:test_trajs}}
% \vspace{-1.25em}
% \end{figure*}

\begin{figure*}[t]
    \begin{minipage}[!ht]{0.25\textwidth}
        \centering
        \subfigure{\includegraphics[width=0.45\textwidth, trim=40 0 20 110,
        clip]{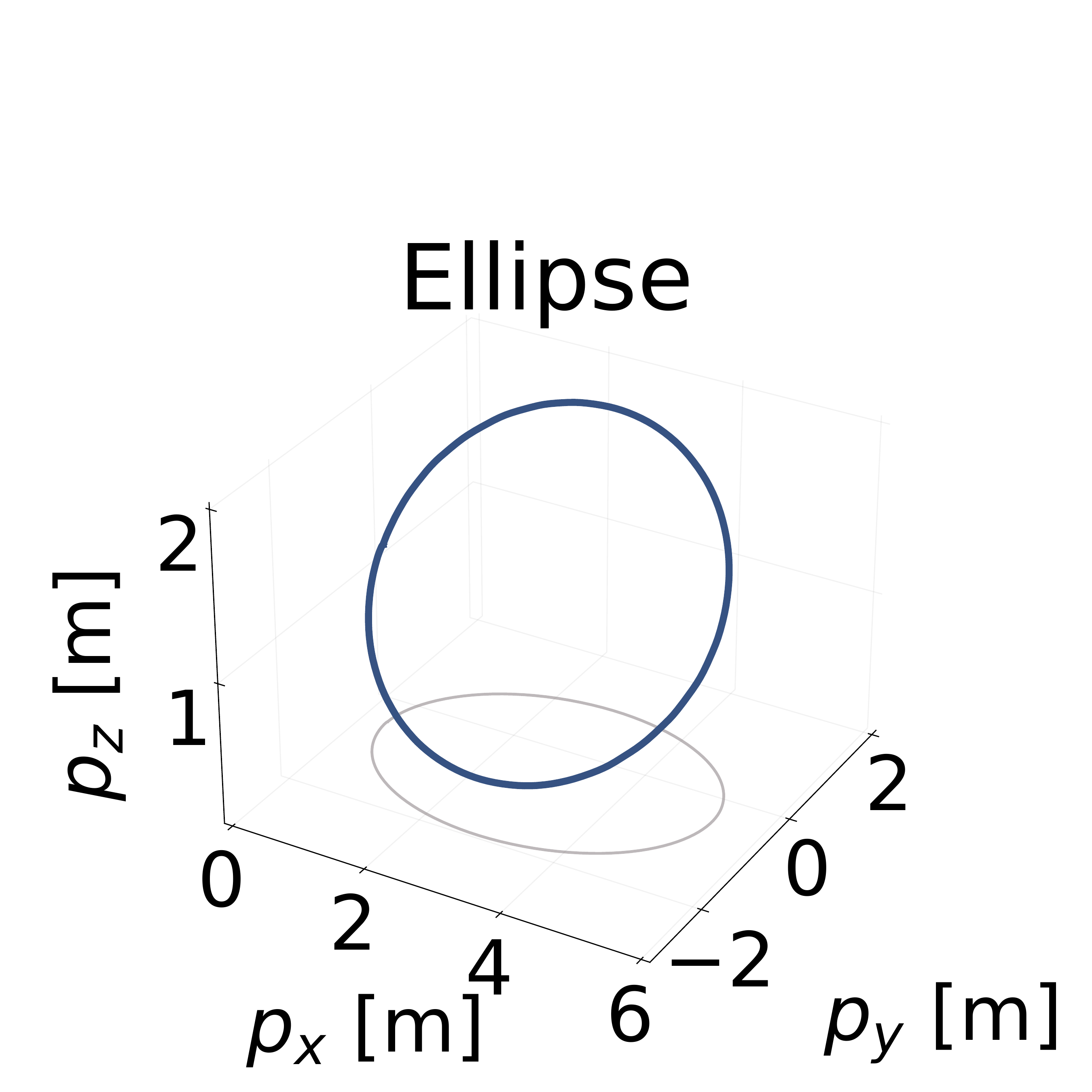}}
        \subfigure{\includegraphics[width=0.45\textwidth, trim=40 0 20 110,
        clip]{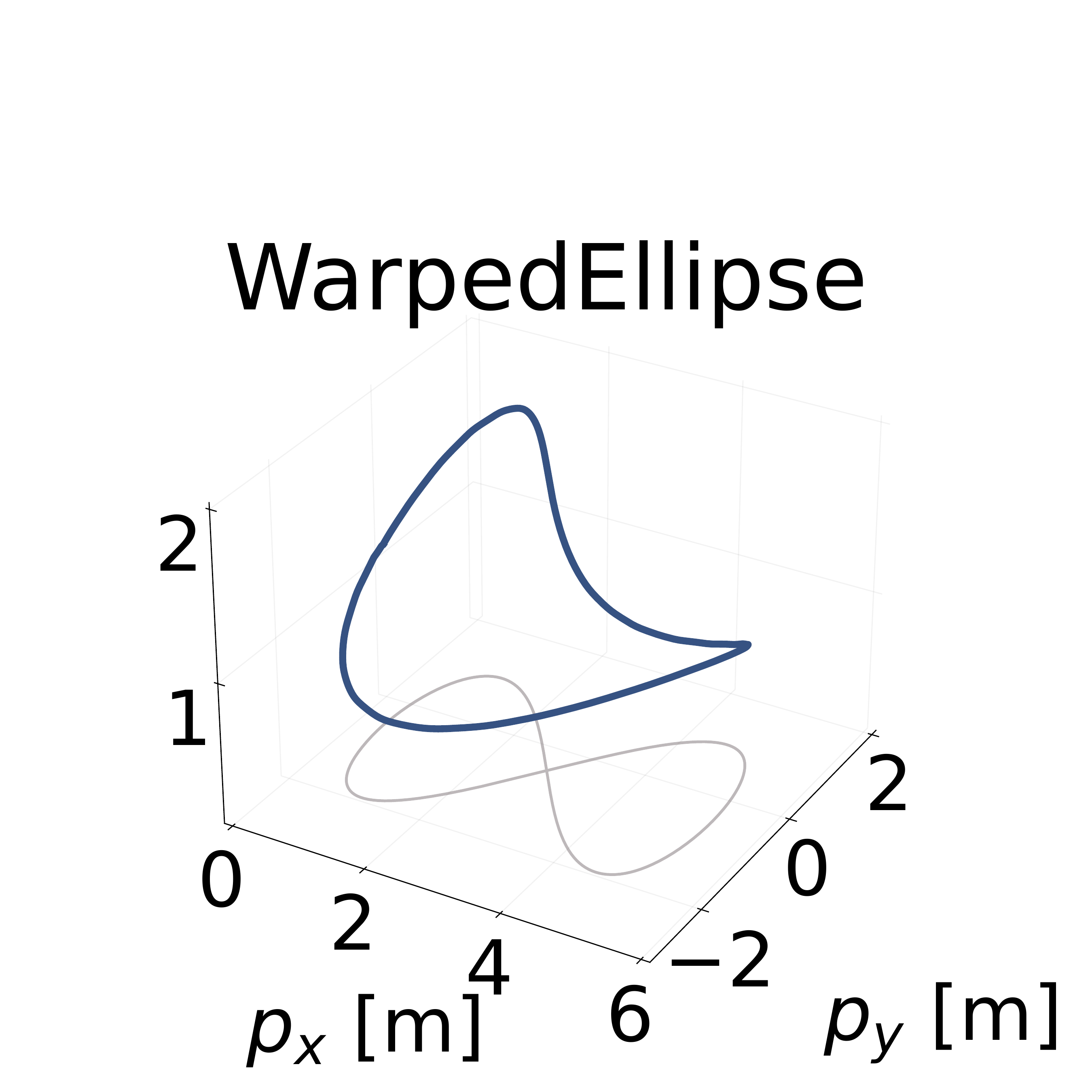}}
        \subfigure{\includegraphics[width=0.45\textwidth, trim=40 0 20 110, clip]{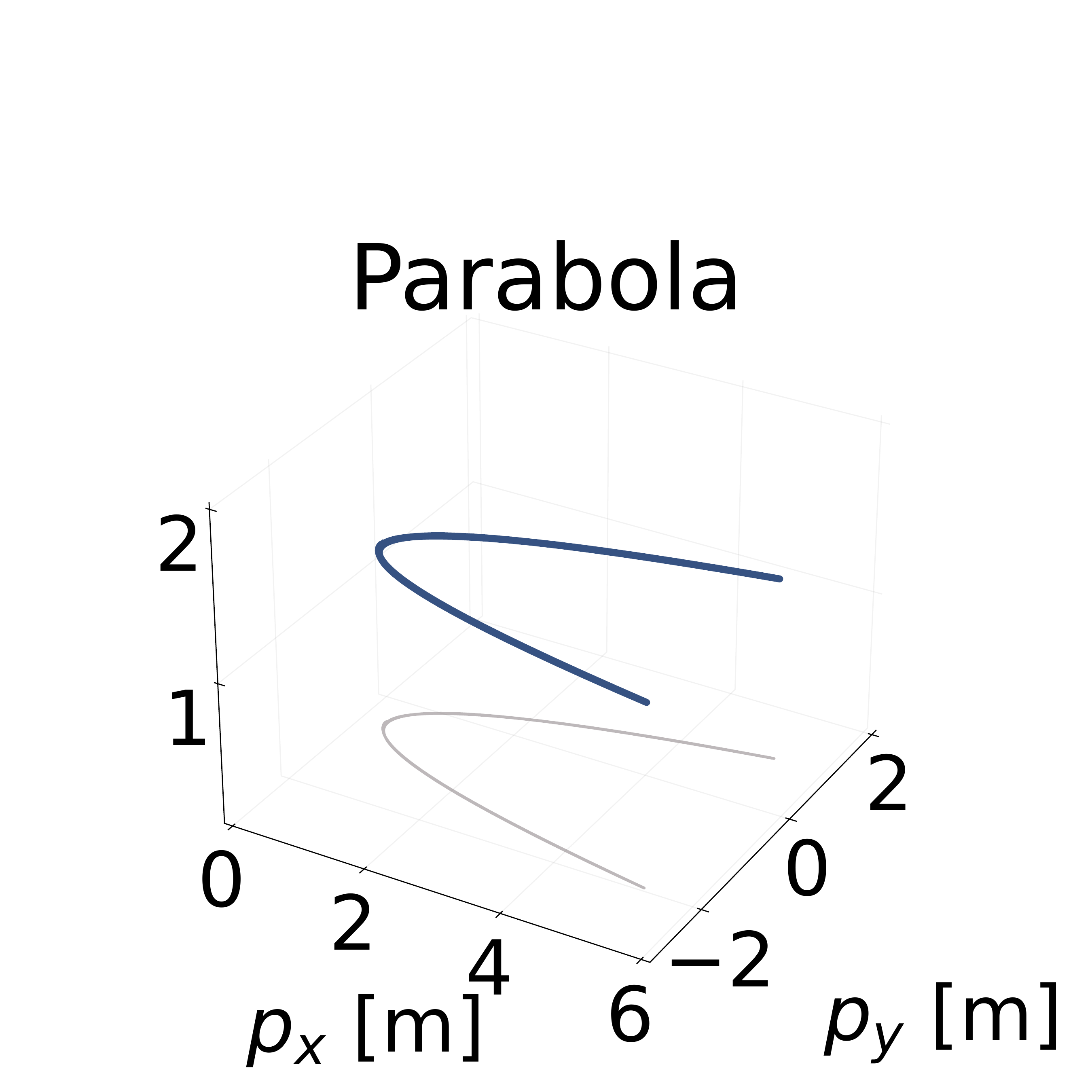}}
        \subfigure{\includegraphics[width=0.45\textwidth, trim=40 0 20 110, clip]{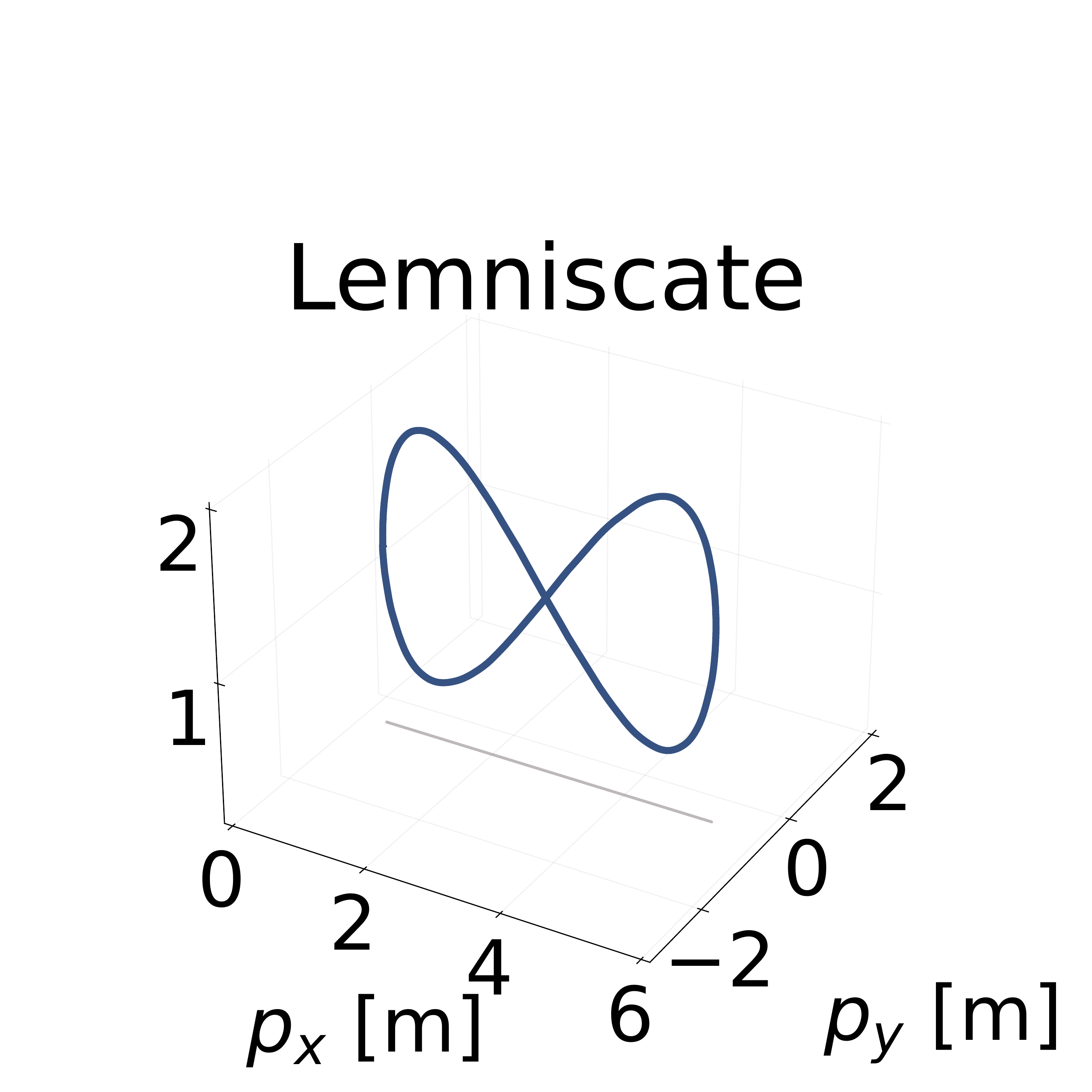}}
        \subfigure{\includegraphics[width=0.45\textwidth, trim=40 0 20 110, clip]{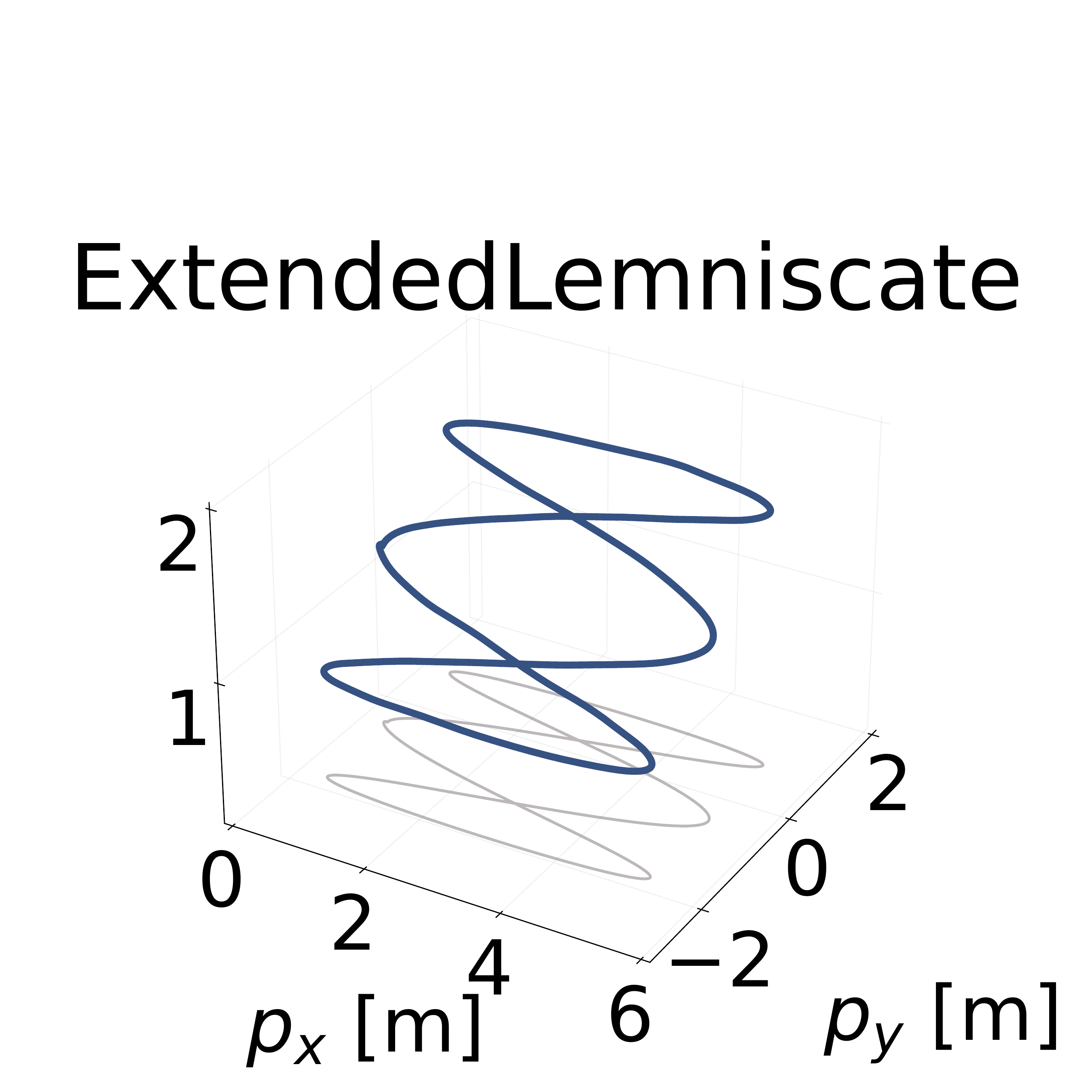}}
        \subfigure{\includegraphics[width=0.45\textwidth, trim=40 0 20 110, clip]{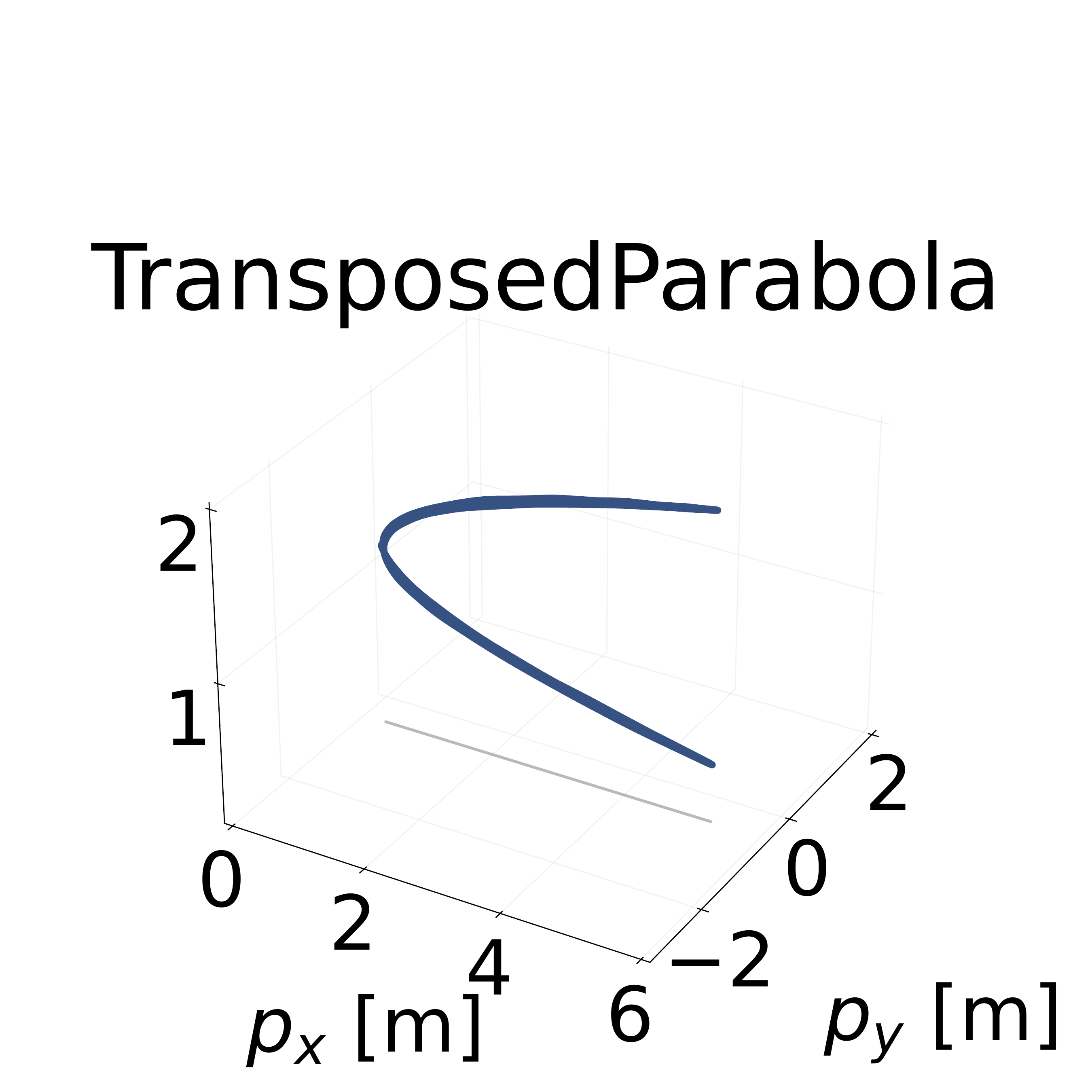}}
        % \vspace{-0.5em}
        \captionof{figure}{Testing trajectories. \label{fig:test_trajs}}
        % \vspace{-1em}
    \end{minipage}
    \hfill
    \begin{minipage}[!ht]{0.75\textwidth}
        \centering
        \includegraphics[width=\textwidth, trim=20 20 70 0, clip] {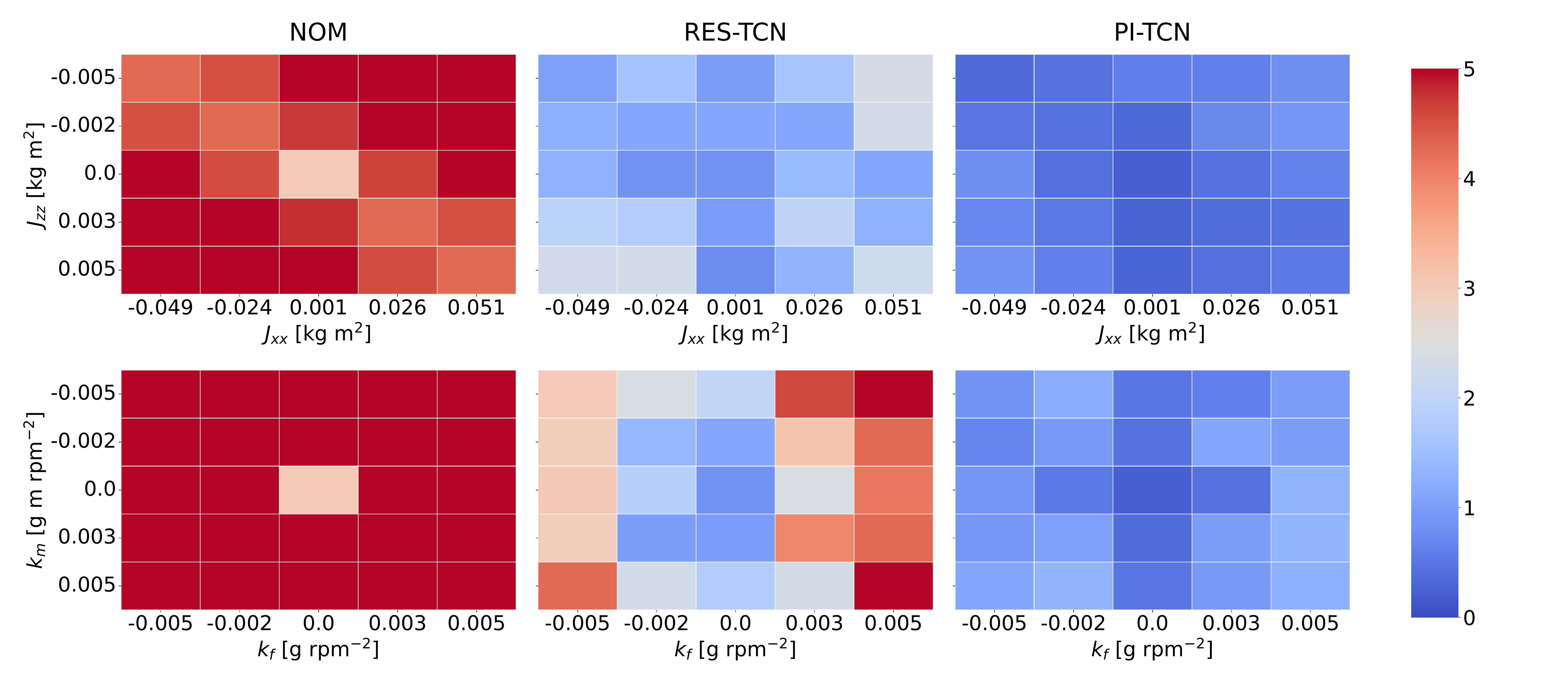}
        \vspace{-0.8em}
        \captionof{figure}{Predictive performance analysis when using different physical priors. \label{fig:heatmaps}}
    \end{minipage}
    \vspace{-0.75em}
\end{figure*}

\subsection{Predictive Performance}
\label{sec:predictiveperf}
We compare the predictive performance of PI-TCN and the baselines on unseen trajectories collected in the real world. The trajectories' speeds and accelerations cover the entire performance envelope of our quadrotor's platform, making the controller highly sensitive to model inaccuracies. For these experiments, we use the root mean squared error (RMSE) between ground truth and predicted accelerations.

Table~\ref{tab:prediction_error} reports the predictive performance in terms of accuracy and computational time of PI-TCN and the baselines, while Figure~\ref{fig:prediction_performance} illustrates the predictive performance over a sample aggressive maneuver (see supplementary material for additional prediction results). 
All the models offer accurate predictive performance over low-speed trajectories, such as Ellipse and WarpedEllipse. However, as speeds and accelerations increase, complex aerodynamic effects acting on the quadrotor's platform significantly degrade the flight performance.
Particularly, NOM is no longer capable of guaranteeing accurate predictions, which may lead to fatal control or navigation failures. Conversely, learning-based approaches demonstrate consistent performance over all maneuvers, capturing all the complex non-linear effects, and performing accurate predictions.
\rebuttal{The results also demonstrate the importance of embedding physical laws as soft constraints in the learned dynamics. PI-TCN consistently outperforms RES-TCN and MSE-TCN despite requiring the same computational cost.
The improved generalization capabilities of PI-TCN may be explained by the fact that the physical constraints can be interpreted as a regularization term that favors well-generalizable solutions lying in large flat valleys of the loss landscape while skipping poorly-generalizable solutions located in sharp regions \cite{chaudhari2017widevalleys}.
Moreover, the results show that encoding the states and control inputs history in a more compact hidden representation improves the accuracy of PI-MLP predictions. Specifically, dilated convolutional layers better capture time-correlated features than dense layers.
Finally, learning-based approaches showcase high generalization capabilities when trained in simulation and directly deployed on unseen real-world data. Even though the performance of these models lightly degrades during the domain shift, they are still significantly more accurate than NOM.
}
The drawback of learning-based approaches is the computational time required to generate the predictions. Even though PI-TCN makes more accurate predictions with respect to the simpler \rebuttal{PI-MLP}, this comes at an $8$x time cost.

\begin{figure}[H]
    \centering
    \subfigure{\includegraphics[width=\linewidth, trim=70 0 80 10, clip]{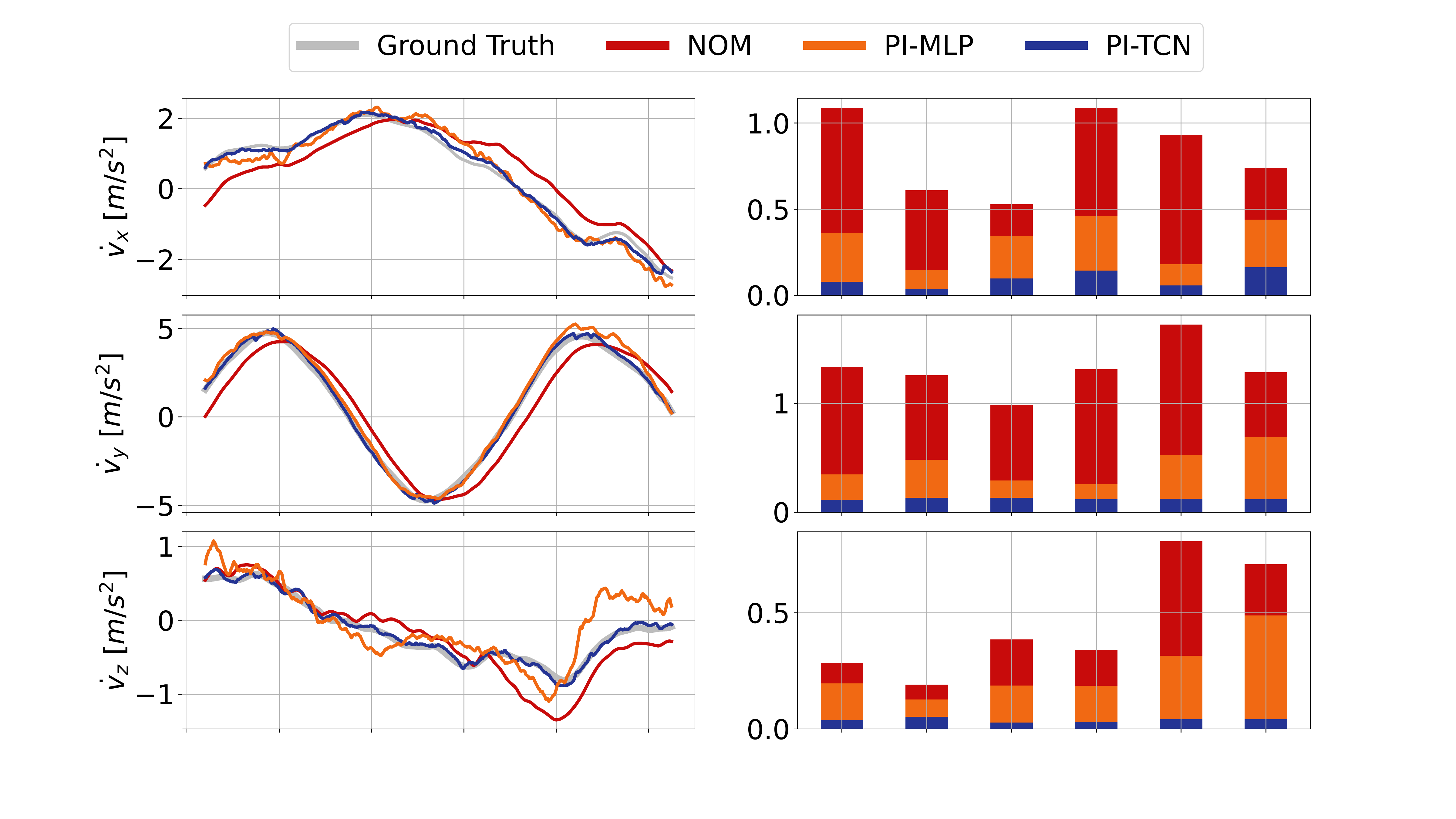}}
\end{figure}
\vspace{-4.15em}
\begin{figure}[H]
    \centering
    \subfigure{\includegraphics[width=\linewidth, trim=70 10 80 90, clip]{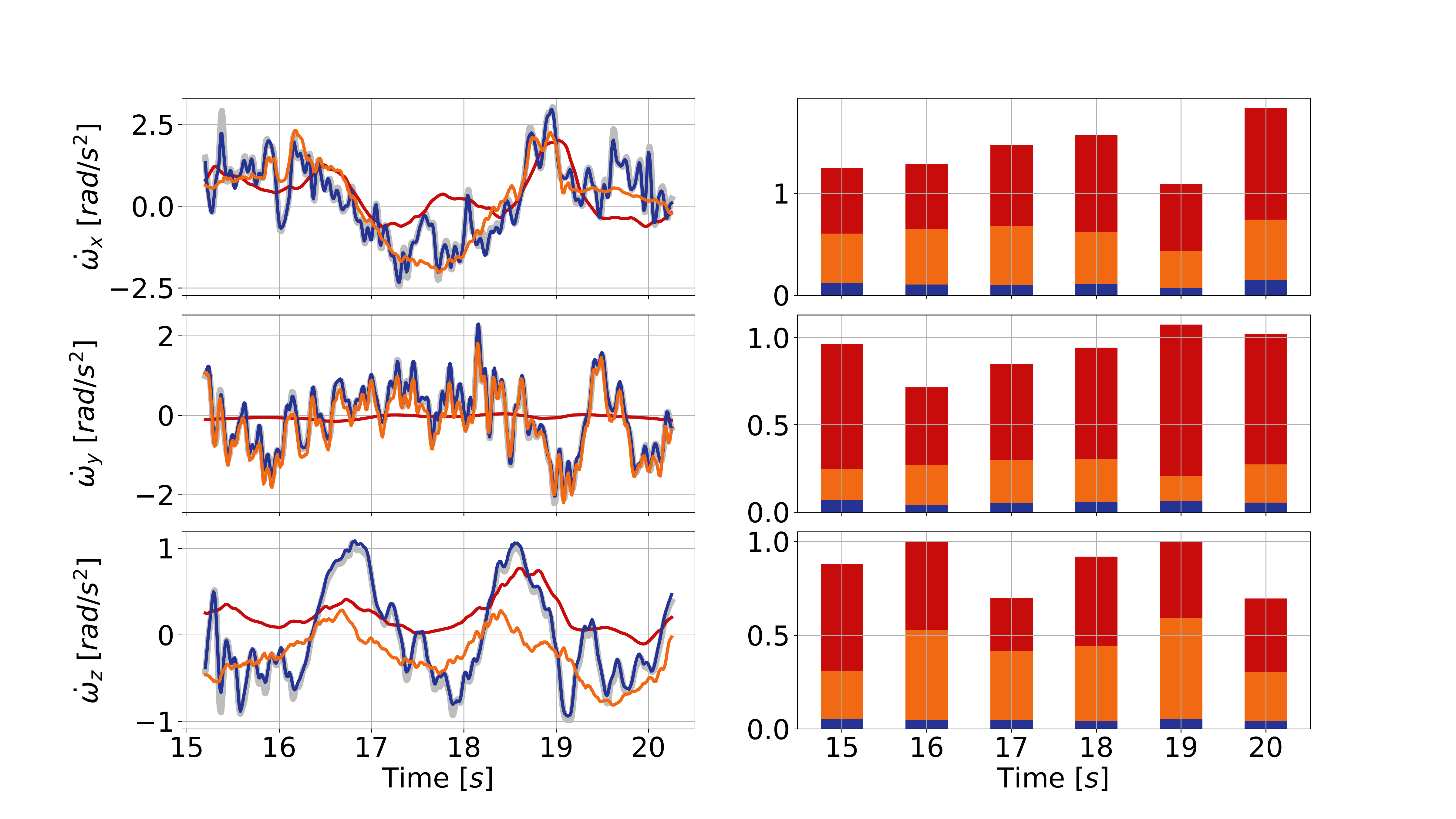}}
    \vspace{-1.75em}
    \caption{Predictive performance over WarpedEllipse\_1. \textbf{Left:} prediction over a trajectory slice. \textbf{Right:} RMSE between prediction and ground truth (stacked with no overlap). \label{fig:prediction_performance}}
\end{figure}

\rebuttal{
\subsection{Robustness to Defective Physical Prior}
\label{sec:robustness2lambda}
The predictive performance of PI-TCN is controlled by the introduction of the physics-inspired loss during the training process. This loss is regulated by the $\lambda$ hyper-parameter which reflects how confident we are in the physical constraints of our system, i.e. how much we want to trade-off between the exploitation of the available nominal model and exploration of the loss landscape.
If our confidence is well-placed, i.e. an accurate physical prior is available and $\lambda > 0$, the predictive performance of the dynamical model is significantly improved, as demonstrated by Table~\ref{tab:prediction_error}.
However, we may be erroneously overconfident about the available physical prior. In such a scenario, including the physics-inspired loss would inevitably degrade the predictive performance of the trained model.
In this section, we conduct an in-depth analysis of the predictive performance of NOM, RES-TCN, and PI-TCN models when different physical priors are available.
Specifically, we first perturb the diagonal moment of inertia matrix $J_{xx}, J_{zz}$ and the rotor thrust and torque constants $k_f, k_\tau$ and then evaluate the models' predictive performance over the testing data. Note that RES-TCN and PI-TCN had to be trained from scratch for each combination of the physical parameters. 

Figure~\ref{fig:heatmaps} illustrates the results of these experiments. Each cell of the heatmaps corresponds to a different physical prior configuration and its color intensity corresponds to the RMSE between the predicted and ground-truth accelerations. Each heatmap central cell coincides with the physical prior configuration adopted in Table~\ref{tab:prediction_error}.
The predictive performance of NOM is strongly affected by the choice of the parameters, resulting in poor prediction results when introducing even small perturbations to the physical prior.
Contrarily, RES-TCN benefits from the expressive power of the temporal convolutional network to balance the degraded physical prior, demonstrating accurate predictive performance with relatively small perturbations. However, as the perturbations increase, RES-TCN can no longer guarantee accurate predictions. In fact, by learning only the residual dynamics, the model does not explore sufficiently the loss landscape and the learned term is not sufficient to balance the physical prior inaccuracy.
This limitation is overcome by PI-TCN which embeds the physical prior only as a soft constraint to direct the training process. The resulting learned dynamical model is more robust to prior changes and the predictive performance experiences only a minor performance reduction.
}

\subsection{Closed-Loop Tracking Performance}
We validate the learned dynamical model against the nominal model in the real-world setting. Specifically, we employ the MPC formulated in Section~\ref{sec:mpc} to control our quadrotor to track multiple trajectories with different models. We compare the tracking performance based on the positional RMSE.
Due to the computational cost imposed by the temporal convolutional network on the controller optimization, we perform the tracking task using the \rebuttal{PI-MLP} model. Moreover, to make the comparison fair with NOM in terms of available information, we set $T=1$.

Table~\ref{tab:tracking_error} reports the tracking performance results on the tested trajectories.
The NOM model captures some gross dynamics that allow tracking the reference trajectory up to some degree. However, the tracking performance degrades significantly \rebuttal{for more aggressive trajectories}, in particular when the angular acceleration mismatch between the nominal prediction and the ground truth is more relevant (e.g., sharp turns), \rebuttal{as for Lemniscate, ExtendedLemniscate, and TrasposedParabola}.
\rebuttal{Conversely, PI-TCN better captures the highly nonlinear angular accelerations and this results in an improved flight performance with a positional RMSE decrease by up to $\times2.7$.}
Generally, over the entire set of test trajectories, the learning model \rebuttal{consistently and significantly outperforms the nominal dynamics}. This empirically demonstrates that the learned model can extract the system's dynamics structured in the data better than the simpler nominal model.

\rebuttal{Figure~\ref{fig:tracking_error} illustrates the tracking performance on the Lemniscate trajectory.
By leveraging PI-TCN dynamical model, the MPC can reach higher linear and angular accelerations (as illustrated by the histograms) while still guaranteeing stable control.
Consequently, the tracking performance using PI-TCN improves the positional RMSE error by over $50\%$ compared to NOM.
}

\begin{table}[t]
    \centering
    \caption{\label{tab:tracking_error}}
    \vspace{-0.75em}
    \caption*{\scshape Tracking Performance in Real-World}
    \rebuttal{
    \vspace{-0.3em}
    \begin{tabular}{c c c}
    \toprule \toprule
    Trajectory & Nominal & Ours\\
    \midrule
    Ellipse\_1 & $ 0.061 \pm 0.001 $ & $ \textbf{0.059} \pm \textbf{0.002} $\\
    Ellipse\_2 & $ 0.126 \pm 0.011 $ & $ \textbf{0.088} \pm \textbf{0.023} $\\
    WarpedEllipse\_1 & $ 0.051 \pm 0.012 $ & $\textbf{0.044} \pm \textbf{0.009} $\\
    WarpedEllipse\_2 & $ 0.098 \pm 0.014 $ & $\textbf{0.069} \pm \textbf{0.017} $\\
    Lemniscate & $ 0.199 \pm 0.032 $ & $\textbf{0.098} \pm \textbf{0.011} $\\
    ExtendedLemniscate & $ 0.272 \pm 0.033 $ & $\textbf{0.101} \pm \textbf{0.019} $\\
    Parabola & $ 0.111 \pm 0.006 $ & $ \textbf{0.092} \pm \textbf{0.021} $\\
    TrasposedParabola & $ 0.322 \pm 0.049 $ & $\textbf{0.162} \pm \textbf{0.051} $\\
    \bottomrule \bottomrule
    \end{tabular}
    % \vspace{-0.5em}
    }
\end{table}

\begin{table}[b]
    % \vspace{-1em}
    \centering
    \caption{\label{tab:ablation_studies}}
    \vspace{-0.75em}
    \caption*{\scshape Ablation Studies}
    \rebuttal{
    \vspace{-0.3em}
    \begin{tabular}{c c c}
    \toprule \toprule
    History & PI-loss & RMSE\\
    \midrule
    \cmark & \cmark & $ 0.22 \pm 0.15 $\\
    \cmark & \xmark & $ 0.55 \pm 0.44 $\\
    \xmark & \cmark & $ 3.36 \pm 1.21 $\\
    \xmark & \xmark & $ 3.52 \pm 1.58 $\\
    \bottomrule \bottomrule
    \end{tabular}
    }
\end{table}

\subsection{Ablation Studies} \label{subsec:ablation_studies}
PI-TCN is based on several components that are designed to improve its predictive performance and generalization capabilities. We validate our design with an ablation study to evaluate the roles of the different network components. In particular, we ablate the following components: (i) the importance of extracting a history of past flight states and control inputs, and (ii) the improved training performance induced by the physics-inspired loss.

Table~\ref{tab:ablation_studies} shows that every component is important, but some of them have a larger impact than others. Specifically, the network trained with the combination of physics-inspired and mean squared error losses better generalizes outside its training set.
However, the most important contribution to the network’s predictive performance is the history of past states and control inputs. Without this component, the network increases its sensitivity to noise in the data and thus its predictive performance degrades significantly.

\section{Discussion and Conclusions}
In this work, we proposed PI-TCN, a deep neural network that extracts quadrotor's dynamics purely from data by leveraging the expressive power of temporal convolutional networks and the generalizability offered by instilling physics laws in the training process. Furthermore, we showed how to exploit the network's predictive performances for accurate predictive trajectory tracking.
The proposed learning method provides several demonstrated advantages over existing methods in the literature.
While classic approaches only rely on present information to estimate the system's dynamics, PI-TCN takes advantage of a history of states and control inputs to capture time-dependent features that would otherwise remain hidden. Moreover, extending the present with past information makes the model less subject to noise in the data.
We demonstrated these capabilities in several experiments where PI-TCN performs accurate predictions both in simulation and real-world settings, consistently outperforming the classical nominal model and the learning-based baselines.
%Moreover, we empirically prove that our network generalizes well when trained in simulation and tested on real-world data, showcasing its predictive capabilities on data outside the training distribution. 
%
While other learning-based methods decouple linear and angular accelerations or only estimate the former, jointly learning linear and angular accelerations allows capturing the hidden dependencies that bound forces and torques for nonholonomic and underactuated systems like the quadrotor. This advantage is also empirically demonstrated by embedding the learned-based model in an MPC framework and accurately tracking trajectories in different flight regimes.

One limitation of the proposed approach is the computational time required by the MPC to solve its online optimization problem. Future works will improve the efficiency of our implementation and leverage GPU parallel computation to fully exploit PI-TCN in the controller horizon.
%
% Finally, we plan to improve our network capabilities of extracting time-dependent features from the history of states and control inputs by varying the history sequence length during training.
%
\rebuttal{
Finally, we plan to use the proposed MPC framework, leveraging the learned dynamical model, as a privileged expert to teach a control policy even more agile maneuvers, such as stunts, flying through narrow windows and thrown hoops \cite{kaufmann2020deepacrob, loianno2017aggressmaneuver}.
}

% references section
\bibliography{IEEEabrv,ref}
\bibliographystyle{IEEEtran}

\clearpage

\section*{Supplementary Material}

\subsection{Collected Data}\label{app:collected_data}
We collect the data by controlling the quadrotor in a series of flights both in simulation and in the real world, resulting in two datasets with analogous trajectories. Each dataset consists of $68$ trajectories.
% %
% Figure~\ref{fig:collected_data} illustrates the Lemniscate trajectory used for testing. The trajectory consists of $6$ lemniscate loops, each covering an area of $5\times1.5$ $\SI{}{\meter}$ on the x-z plane, with speeds ranging from $- 4.26$ to $4.30$ $\SI{}{\meter\per\second}$, linear accelerations from $− 13.14$ to $7.08$ $\SI{}{\meter\per\second\squared}$, and motor speeds picking at $16628$ $\si{rpm}$.
%
We have empirically found that scaling the motor speeds by a factor equal to $0.001$ helps the network's predictive accuracy. In future work, we will improve the preprocessing step of the data by adopting more general assumptions. Moreover, we will aim to improve the filtering of the data, with a focus on angular accelerations.

\rebuttal{Collecting real-world data for training learning models is rather simple because the procedure does not involve any expensive manual labeling (i.e., self-supervised task). The practitioner should manually control the system and record the history sequence of state estimations and control actions. The state estimation can come from any source, such as Vicon or visual-inertial odometry. In the latter case, the state estimates may be noisy and this should be taken into account during training. Training with noisy labels is a well-known problem in machine learning \cite{song2022survey} and represents an exciting future direction for extending this work.}

\rebuttal{We control the quadrotor by using the model predictive controller formulated in Section~III-D with the nominal model. We use this control framework because the data collection is independent of the chosen controller. The data consists of sequences of state estimations and control actions applied on the platform. We are interested in understanding from a given state and by applying a specific control action, which states the quadrotor will reach. This mapping is not affected by the specific control framework whose only task is to predict the next action to apply. Consequently, one can use any control framework to maneuver the quadrotor.}

\begin{table}[b]
    \vspace{-1em}
    \centering
    \caption{\label{tab:nominal_model_params}}
    \vspace{-0.75em}
    \caption*{\scshape Nominal Model Parameters}
    \rebuttal{
    \begin{tabular}{c l l}
    \toprule \toprule
    $m$ & $[\SI{}{\kilo\gram}]$ & $ 0.25 $\\
    $k_f$ & $[\SI{}{\gram}$ $\SI{}{\mathrm{rpm}^{-2}}]$ & $ 4.38 \cdot 10^{-9} $\\
    $k_m$ & $[\SI{}{\gram\meter}$ $\SI{}{\mathrm{rpm}^{-2}}]$ & $ 3.97 \cdot 10^{-11} $\\
    $J_{xx}$ & $[\SI{}{\kilo\gram\meter\squared}]$ & $ 0.000601 $\\
    $J_{yy}$ & $[\SI{}{\kilo\gram\meter\squared}]$ & $ 0.000589 $\\
    $J_{zz}$ & $[\SI{}{\kilo\gram\meter\squared}]$ & $ 0.001076 $\\
    $l$ & $[\SI{}{\meter}]$ & $ 0.076 $\\
    \bottomrule \bottomrule
    \end{tabular}
    }
\end{table}

\subsection{Baselines} \label{app:baselines}
\rebuttal{In this section, we describe the baselines used in this work for validating the predictive performance of PI-TCN, namely NOM, RES-TCN, MSE-TCN, and PI-MLP.

NOM is the nominal model as in Eq.~(1). The model is described by a set of hyper-parameters strictly related to the physical platform, namely the mass $m$, the rotor thrust $k_f$ and torque $k_\tau$ constants, the diagonal moment of inertia matrix $J=diag([J_{xx}, J_{yy}, J_{zz}])$, and the arm length $l$.
We estimate the thrust and torque coefficients by using an RC-benchmark Thrust Stand and we approximate the inertia matrix and arm length through precisely generated CAD models as in \cite{wueest2018estimation}.
Table~\ref{tab:nominal_model_params} summarizes the adopted nominal model parameters.

RES-TCN is the residual model baseline and shares the same architecture as PI-TCN. The key difference between the two is the dynamics that they extract from the training dataset. While PI-TCN fully extracts the dynamics from data using the physical laws only in the loss function as soft constraints, RES-TCN is trained to learn the residual dynamics and uses the physical laws as hard constraints. 
Specifically, the nominal model defined in Eq.~(1) can be reformulated as:
\begin{equation} \label{eq:residual_method}
    \dot{\mathbf{x}} =
    \begin{bmatrix}
    \dot{\mathbf{p}} \\
    \dot{\mathbf{v}} \\
    \dot{\mathbf{q}} \\
    \dot{\boldsymbol{\omega}}
    \end{bmatrix} =
    \begin{bmatrix}
    \mathbf{v} \\
    \frac{1}{m} \biggl[ \mathbf{q} \odot (f + f_{res}) \biggr] + \boldsymbol{g} \\
    \frac{1}{2} ( \mathbf{q} \odot \boldsymbol{\omega} ) \\
    \mathbf{J}^{-1} (\boldsymbol{\tau} + \boldsymbol{\tau}_{res} - \boldsymbol{\omega} \times \mathbf{J} \boldsymbol{\omega})
    \end{bmatrix}.
\end{equation}
where $f_{res}$ and $\boldsymbol{\tau}_{res}$ are the residual forces and torques that are not explained by the nominal model. RES-TCN is trained to extract these residual terms from the collected data.
This implementation is analogous to all residual learning models that consider both translational and rotational residual dynamics, as in \cite{bauersfeld2021neurobem}.

MSE-TCN is the learning model baseline that validates the importance of adding soft physical constraints to improve the generalization performance of the trained model. This model shares the same architecture as PI-TCN. The key difference between the two is the loss function used at training time. Specifically, MSE-TCN is fully trained using $\lambda = 0$.

PI-MLP is the learning model baseline that demonstrates the importance of extracting time-dependent features from a history of states and control actions. This model consists only of the MLP network (i.e., decoder) of PI-TCN.
}

\subsection{Predictive Performance} \label{app:predictions}
We complete the qualitative analysis of the predictive performance of PI-TCN and the baselines on the test trajectory WarpedEllipse\_1. The models are trained in the real-world dataset and have never seen the test trajectory during the training and validation steps.
Figure~\ref{app:pred_perf_warpedellipse_1} and Figure~\ref{app:pred_perf_warpedellipse_2} illustrate the predictive performance of PI-TCN and the baselines. Generally, the predictions of the considered models over the linear accelerations are accurate. PI-TCN is consistently the best performing model, followed by \rebuttal{PI-MLP} and the nominal model. 
\rebuttal{PI-MLP} struggles to generalize well because of the dense architecture that makes it fit the noise in the training data. Conversely, PI-TCN extracts the most important time-dependent features by employing sparse temporal convolutional layers. 
The nominal model makes consistent predictions for the linear accelerations, however, it is always approximating the real dynamics of the system. Thus, by using the nominal model, we may end up having a mediocre flight performance.
The results over the angular accelerations stress more the difference between the considered models.
The nominal model is unable to fit properly the true angular acceleration, particularly over the y-axis. To improve this behavior, we may empirically better fit the nominal model parameters (e.g., inertia, thrust, and torque coefficients) over this trajectory by repeating the system identification task. However, this would result in degrading the performance of the nominal model on other different trajectories. 
MLP predictions over the angular accelerations are coherent with the ones for the linear accelerations. The performance is not accurate but adequate for a safe flight. 
Again, PI-TCN outperforms all the baselines on the angular acceleration predictions by a large margin. The temporal convolutional layers extract the hidden dynamics from past states and control inputs, while the multi-layer perceptron makes accurate predictions over the accelerations.

\subsection{Choosing the History Length} \label{app:history_len}
\begin{figure}[b]
    \centering
    \subfigure{\includegraphics[width=\linewidth, trim=70 10 80 27, clip]{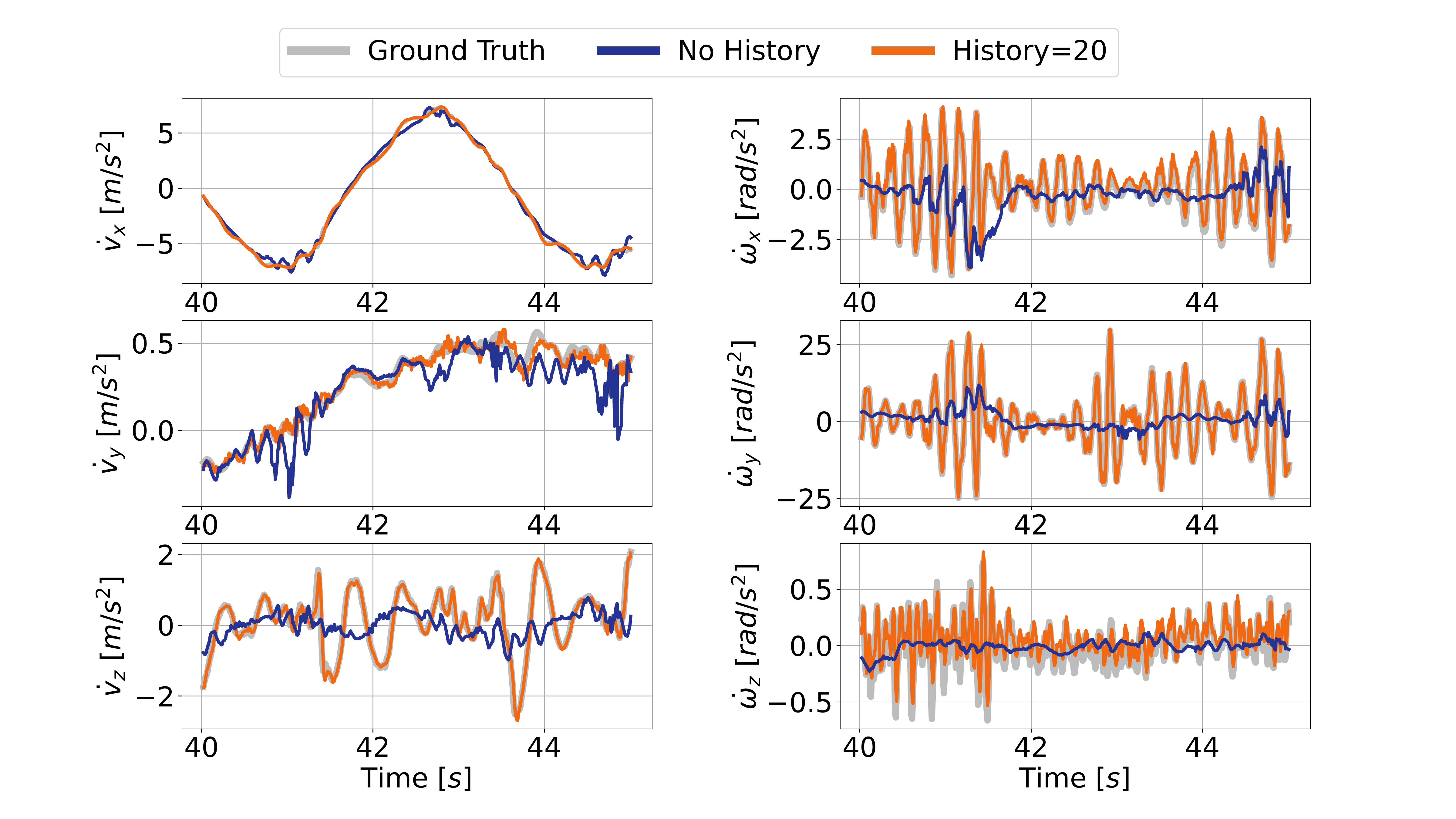}}
    \vspace{-1.5em}
    \caption{Predictive performance over ExtendedLemniscate slice of PI-TCN trained with different history lengths.}
    \label{app:choosing_history_fig}
\end{figure}

Extending the input of the network with a history sequence of past states and control inputs is fundamental for accurately predicting the quadrotor's system dynamics. Thus, we study its effects on the predictive performance of PI-TCN over the test trajectories illustrated in Figure~3.
The network is trained in the real-world dataset with different history lengths and has never seen the test trajectories during the training.
Figure~\ref{app:choosing_history_fig} illustrates the predictive performance of PI-TCN trained with different history lengths, while Table~\ref{tab:choosing_history} reports the results in terms of RMSE between true accelerations and predictions.
Moreover, the results relate the predictive performance with the computational time required by the network to generate the predictions. 
The results demonstrate the importance of extending the input of the network. By employing a unitary history length, the network is too sensitive to noise in the data and struggles to capture the true dynamics of the quadrotor's system. 
On the other hand, by employing a history length of $4$, the network is already capable of achieving accurate performances over the test data. As we increase the history length, the prediction accuracy starts to converge. 
\rebuttal{
In general, we observe that the longer the history length the more accurate the prediction is. This is motivated by the fact that no matter the model we choose, the more data available the better the prediction is.
}
However, at the same time, the computational cost for considering a long history length increases significantly. One should choose the best trade-off between accuracy and computational time based on the considered task and system.

\rebuttal{
\subsection{Choosing $\lambda$}
One of the key contributions of this work is the physics-inspired loss function that embeds soft physical constraints in the network training process and extends the classical mean squared error loss formulation. The physics-inspired and the mean squared error losses in Eq.~(4) are balanced by an hyper-parameter $\lambda$. This hyper-parameter should reflect how confident we are in the physics constraints of our system. If we have access to an unreliable physical model, $\lambda$ should be small to let the neural network fully explore the loss landscape. On the other hand, if we are confident in the available physical model, $\lambda$ can be large to fully exploit the prior.

If $\lambda$ is badly chosen, then the training process would be negatively affected. We have studied the effects of an accurate prior in Section~V-A and the robustness of the proposed approach to different defective priors in Section~V-B. The results show that, when an inaccurate prior is available, the proposed approach still provides satisfactory predictive performance. This can be explained by the fact that during training the network can both exploit the physical prior and explore the loss landscape. Conversely, residual approaches fully exploit the degraded physical prior, thus showing poor performances.

In this work, we train PI-TCN for half the training process by setting $\lambda = 0$. This ensures that the network fully explores the optimization space in a self-supervised fashion. Then, we restart the training using $\lambda = 1$ for the remaining training iterations to stabilize the training process convergence. In future work, we plan to optimize the choice of $\lambda$. Specifically, grid-search algorithms \cite{bergstra2011tuning} or statistical techniques such as Bayesian Optimization \cite{snoek2012bo} can be employed to estimate the best value for $\lambda$ for a given data set.
}

\begin{table}[t]
    \centering
    \caption{\label{tab:choosing_history}}
    \vspace{-0.75em}
    \caption*{\scshape Choosing History Length}
    \rebuttal{
    \begin{tabular}{c c c}
    \toprule \toprule
    History Length & Inference $[\SI{}{\milli\second}]$ & RMSE\\
    \midrule
    $20$ & $1.73$ & $0.23 \pm 0.15$ \\
    $10$ & $1.55$ & $0.24 \pm 0.22$ \\
    $4$ & $1.47$ & $0.82 \pm 0.48$ \\
    $1$ & $1.42$ & $1.32 \pm 1.01$ \\
    $0$ & $1.20$ & $3.36 \pm 1.21$ \\
    \bottomrule \bottomrule
    \end{tabular}
    \vspace{-1em}
    }
\end{table}

\rebuttal{
\subsection{Control Design Setup}
We report the implementation details of the MPC design employed in this work for both collecting the data and validating the proposed learned dynamics.
The MPC formulates an optimization problem that finds a sequence of inputs within a fixed time horizon with $N$ discretized steps by optimizing a given objective function. 
We use a horizon covering the evolution of the system over $1~\si{s}$ and discretize it into $N=20$ steps. We use the high-performance interior-point method solver (HPIPM) \cite{frison2020hpipm} and perform the MPC optimization with a real-time iteration (RTI) \cite{diehl2005rti} scheme. To increase the stability of the controller, we employ the Levenberg-Marquardt regularizer. Using these settings, the MPC runs at frequencies higher than $100~\si{Hz}$ with both the nominal and the learning-based dynamical models, enabling real-time control for practical real-time robot tasks.
}

\begin{figure*}
\centering
\subfigure{\includegraphics[width=0.49\linewidth, trim=70 10 80 20, clip]{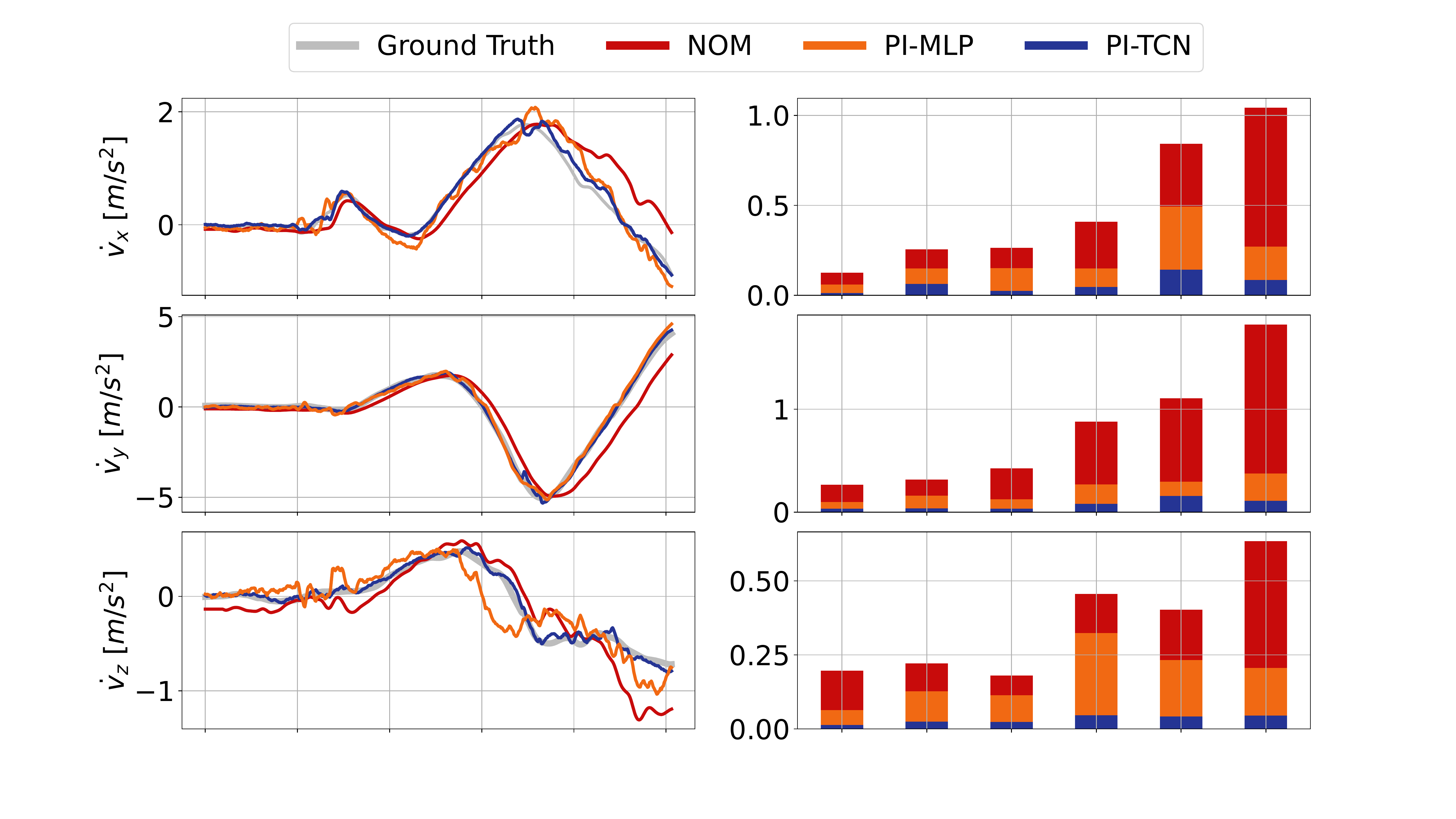}}
\subfigure{\includegraphics[width=0.49\linewidth, trim=70 10 80 20, clip]{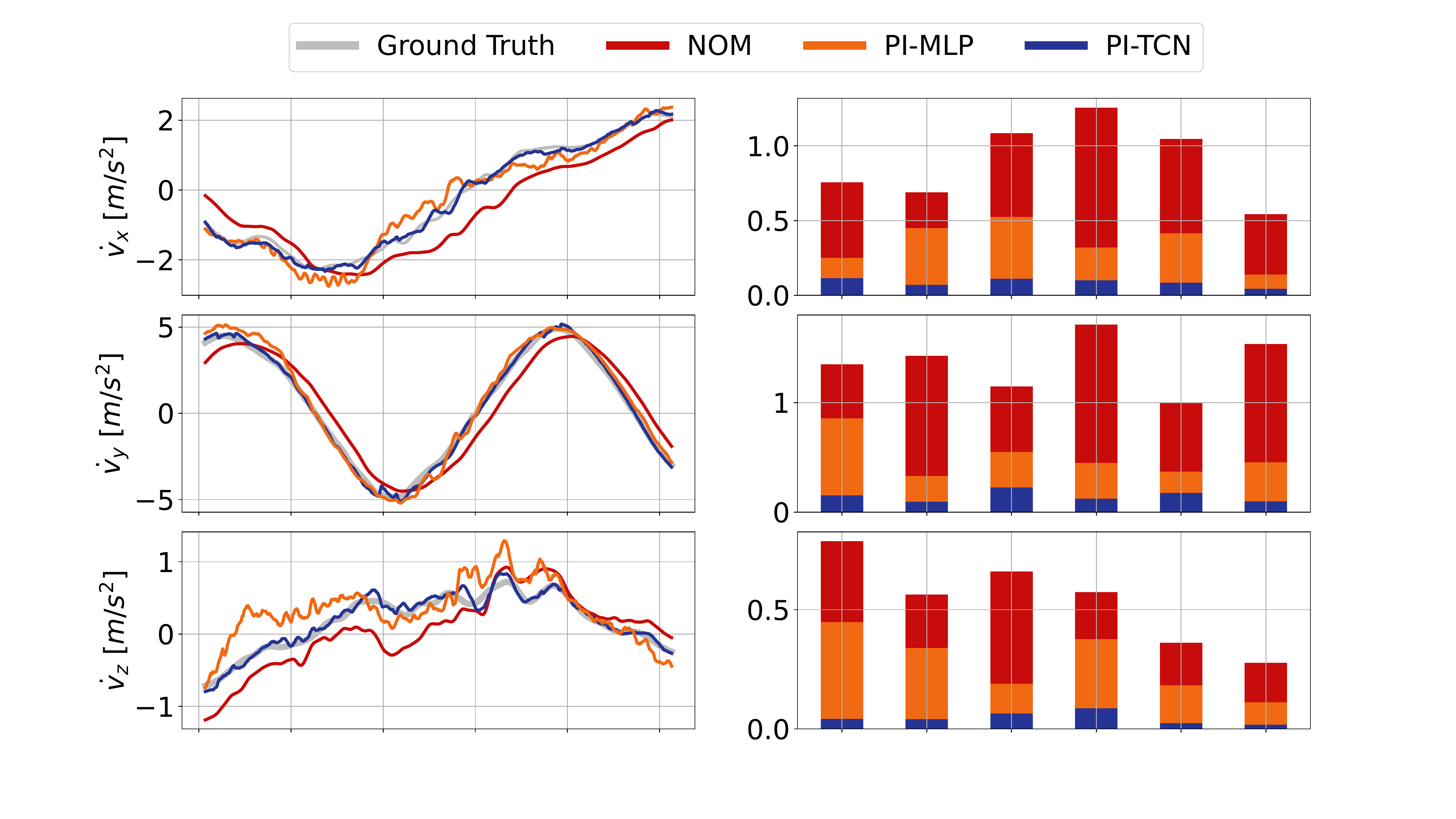}}
\vspace{-5em}
\end{figure*}
\begin{figure*}
\centering
\subfigure{\includegraphics[width=0.49\linewidth, trim=70 10 80 20, clip]{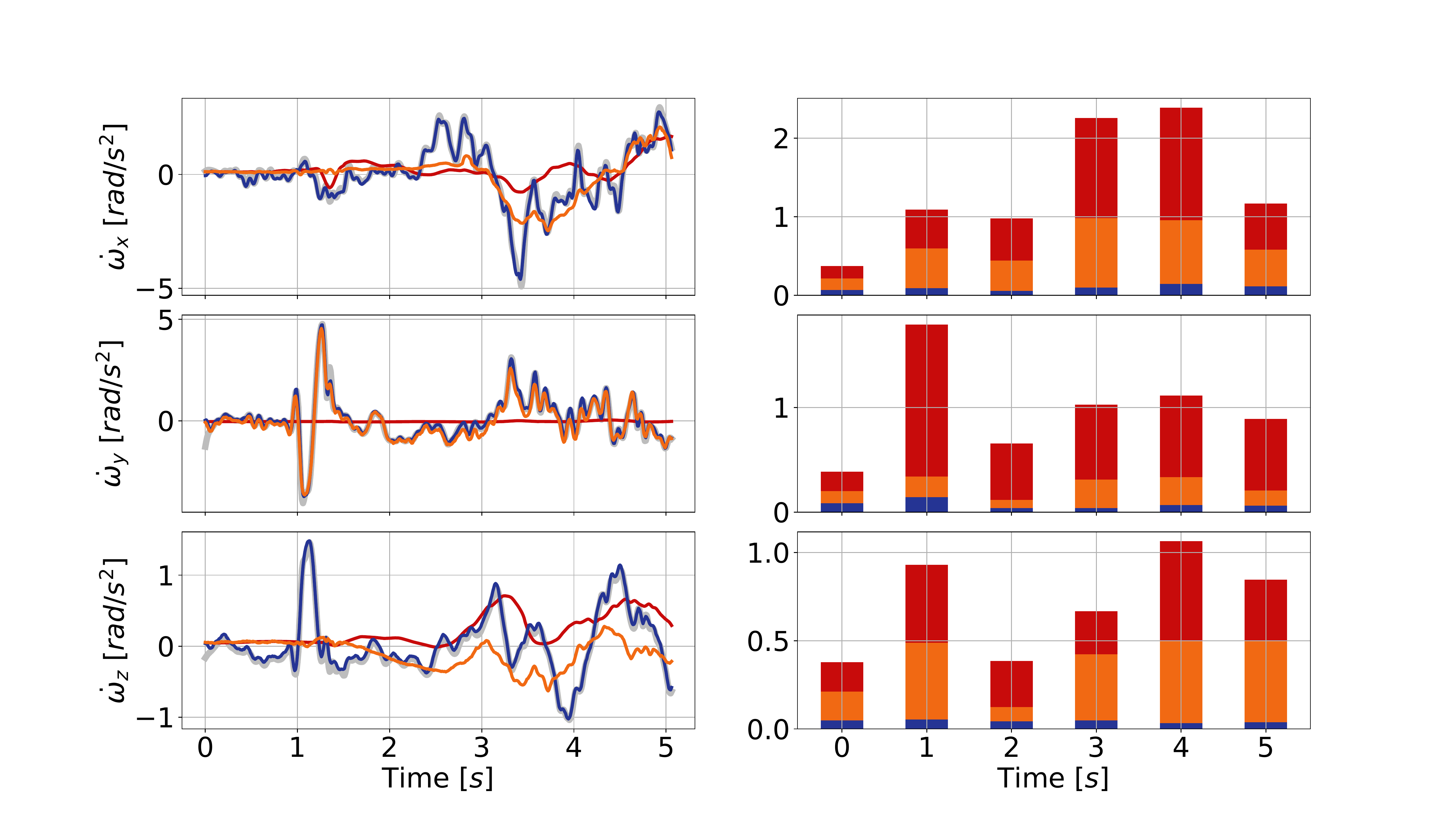}}
\subfigure{\includegraphics[width=0.49\linewidth, trim=70 10 80 20, clip]{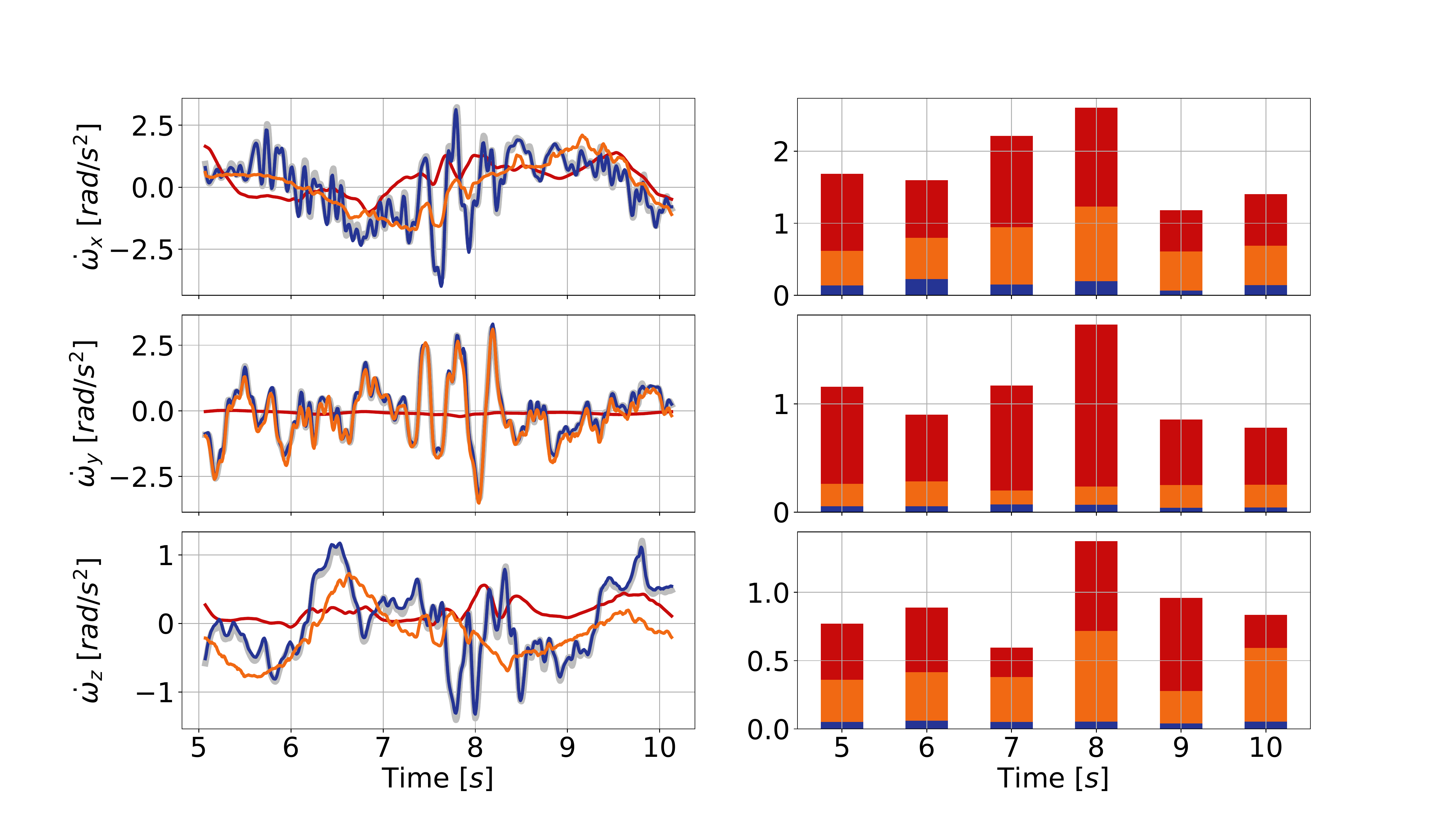}}
\end{figure*}
\begin{figure*}
\centering
\subfigure{\includegraphics[width=0.49\linewidth, trim=70 10 80 20, clip]{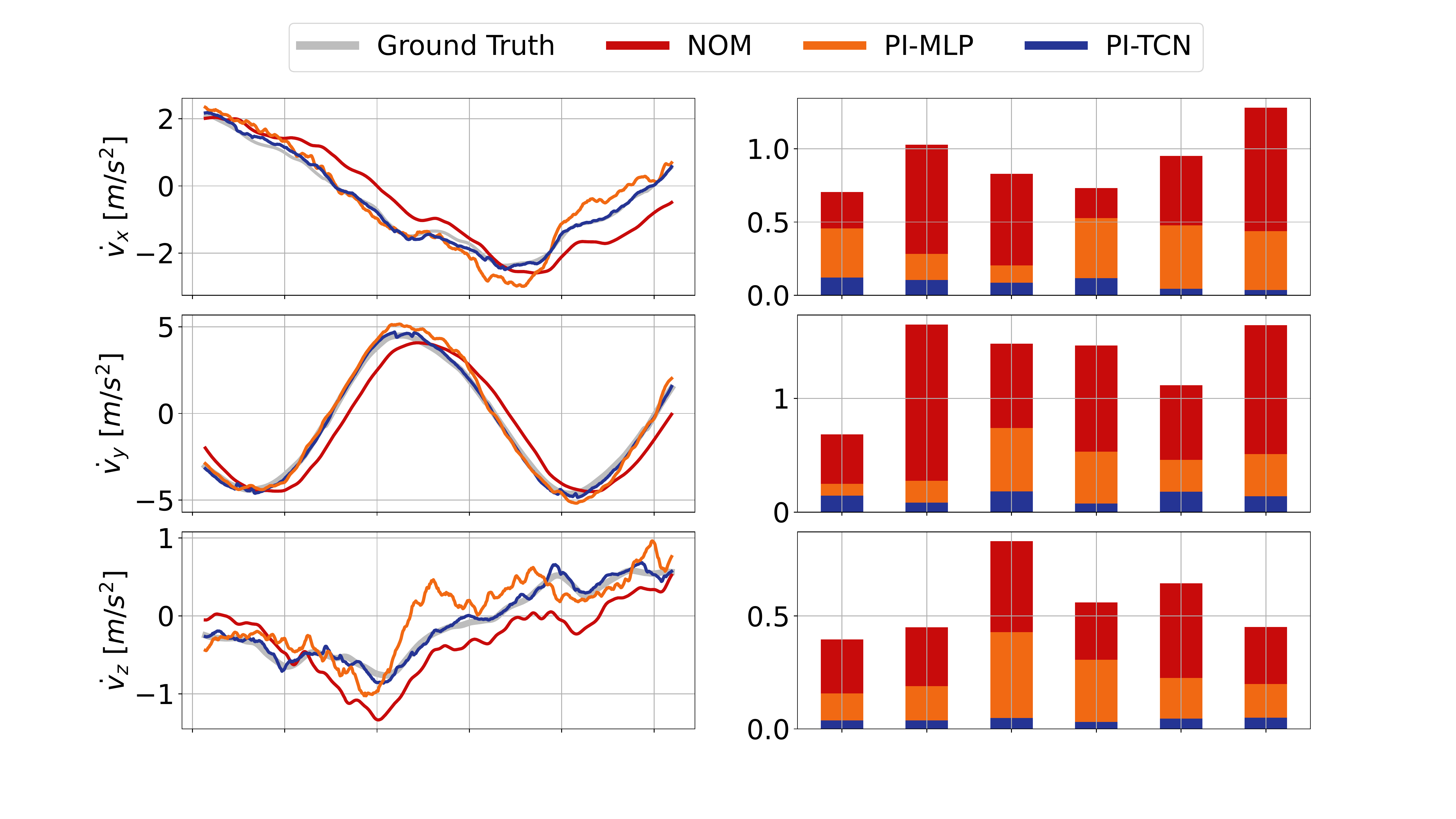}}
\subfigure{\includegraphics[width=0.49\linewidth, trim=70 10 80 20, clip]{images/predictive_performance/WarpedEllipse_7_3.pdf}}
\vspace{-5em}
\end{figure*}
\begin{figure*}
\centering
\subfigure{\includegraphics[width=0.49\linewidth, trim=70 10 80 20, clip]{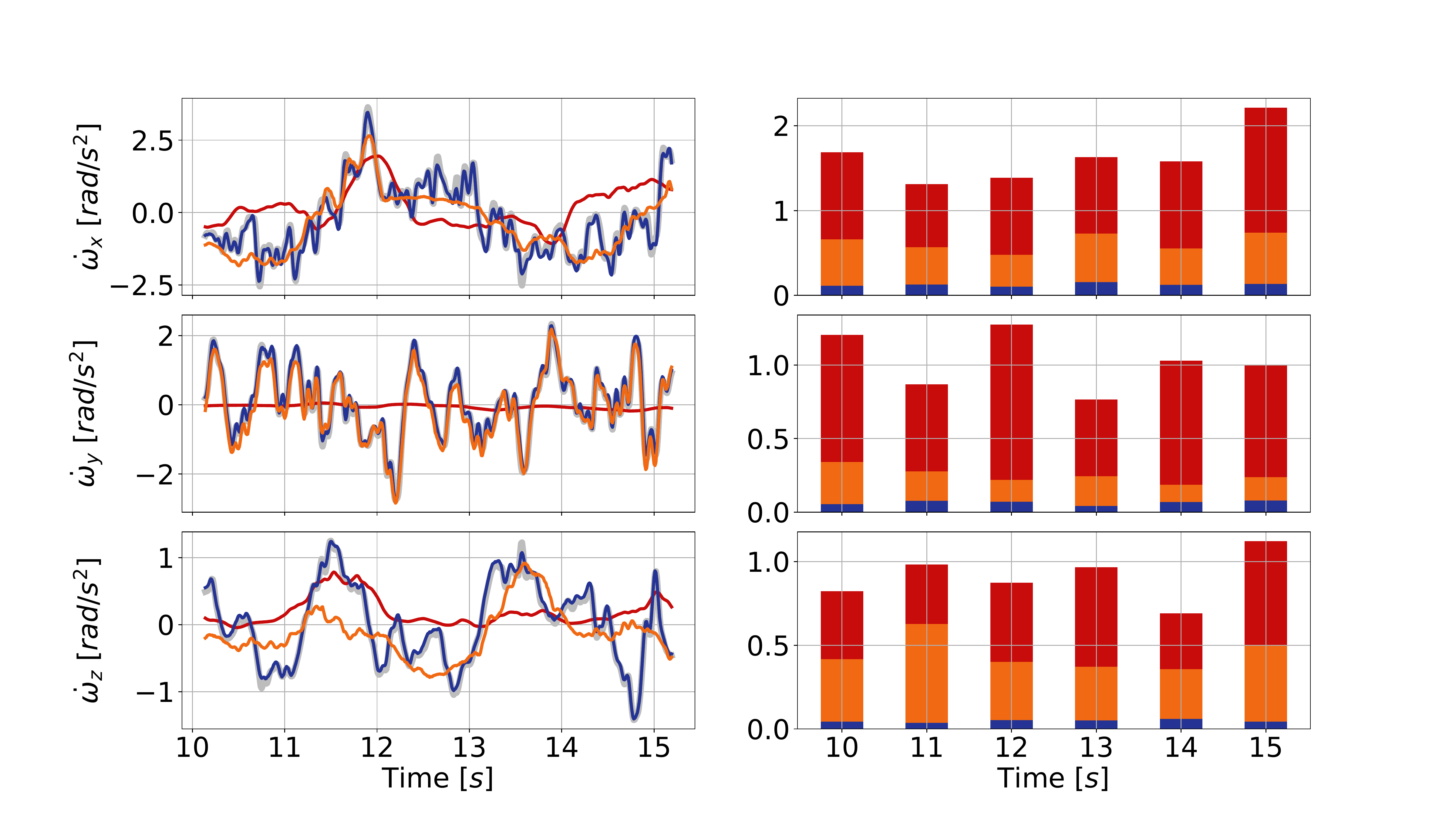}}
\subfigure{\includegraphics[width=0.49\linewidth, trim=70 10 80 20, clip]{images/predictive_performance/WarpedEllipse_7_12.pdf}}
\caption{Predictive performance over the first half of WarpedEllipse\_1. \textbf{Left:} prediction over a slice of the trajectory. \textbf{Right:} root mean square error between prediction and ground truth (stacked vertically with no overlap).}
\label{app:pred_perf_warpedellipse_1}
\end{figure*}

\begin{figure*}
\centering
\subfigure{\includegraphics[width=0.49\linewidth, trim=70 10 80 20, clip]{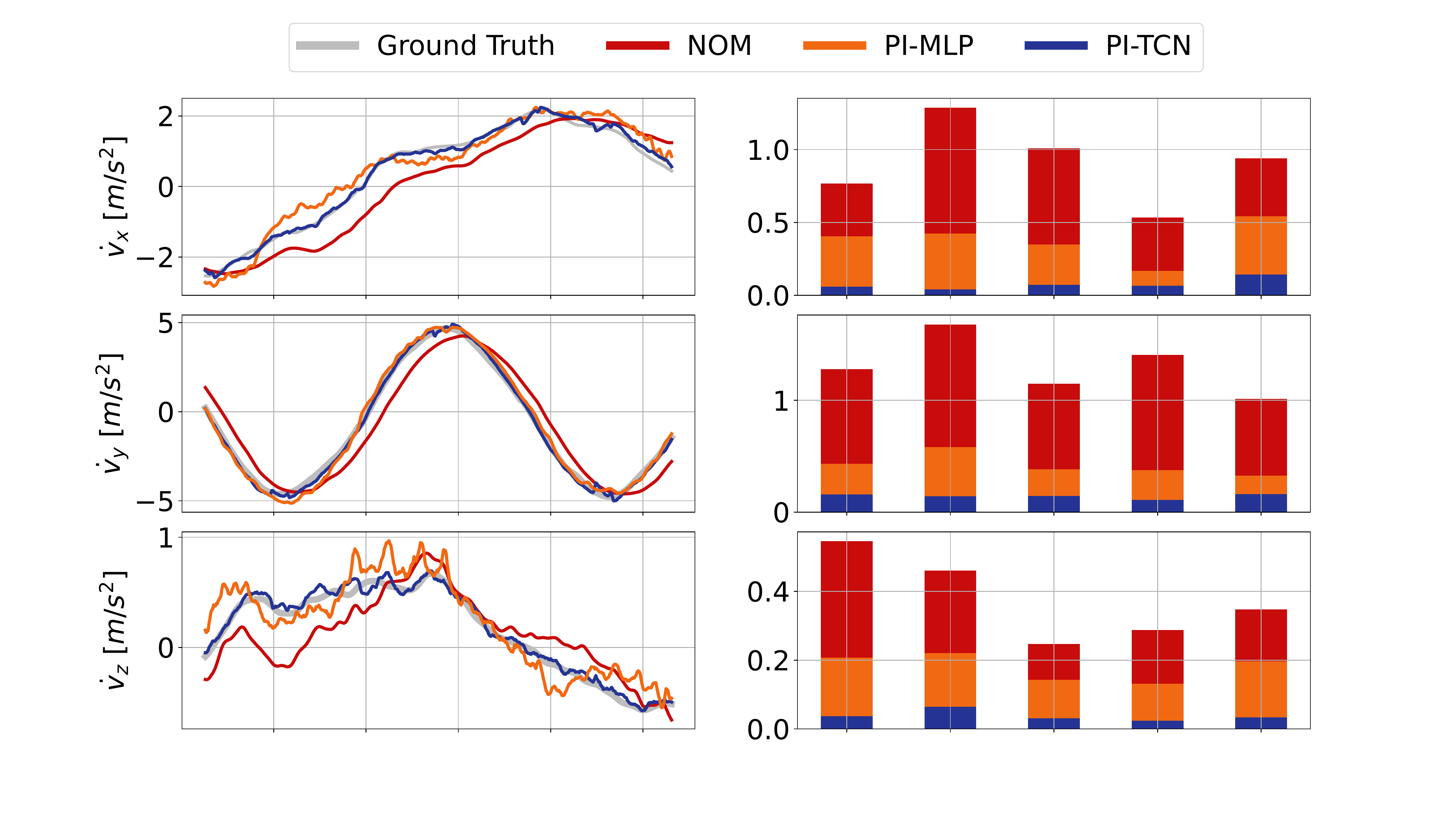}}
\subfigure{\includegraphics[width=0.49\linewidth, trim=70 10 80 20, clip]{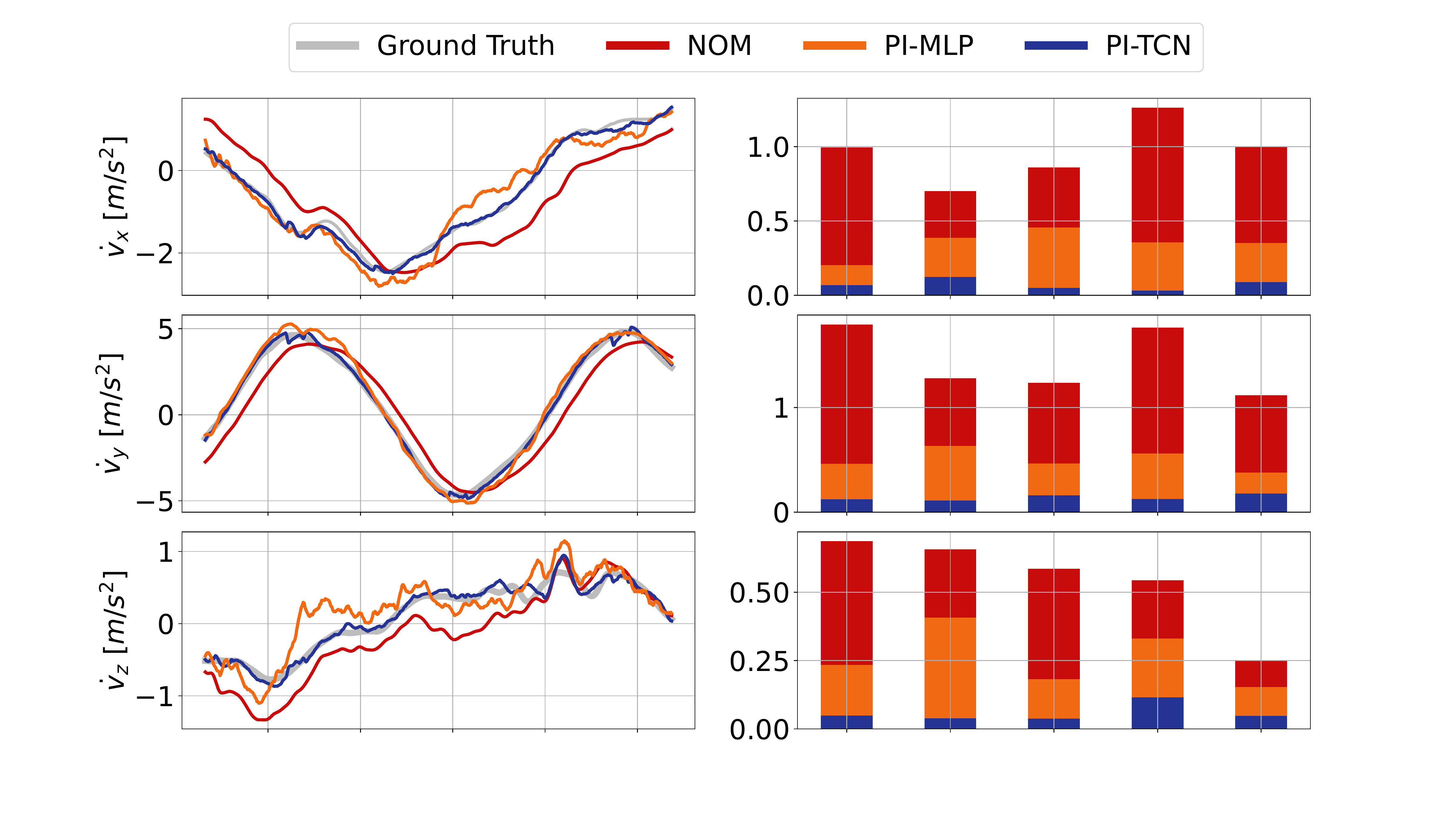}}
\vspace{-5em}
\end{figure*}
\begin{figure*}
\centering
\subfigure{\includegraphics[width=0.49\linewidth, trim=70 10 80 20, clip]{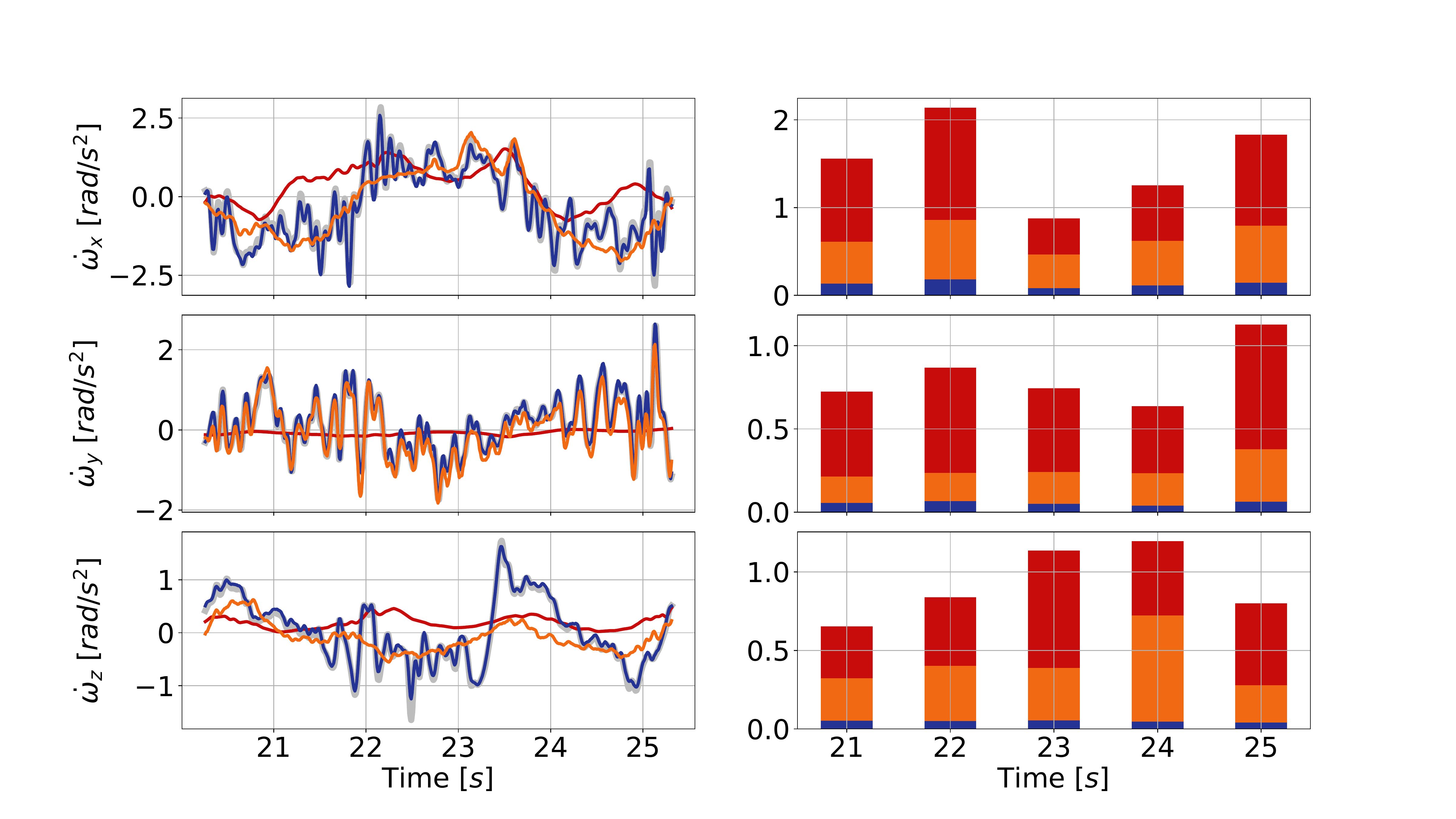}}
\subfigure{\includegraphics[width=0.49\linewidth, trim=70 10 80 20, clip]{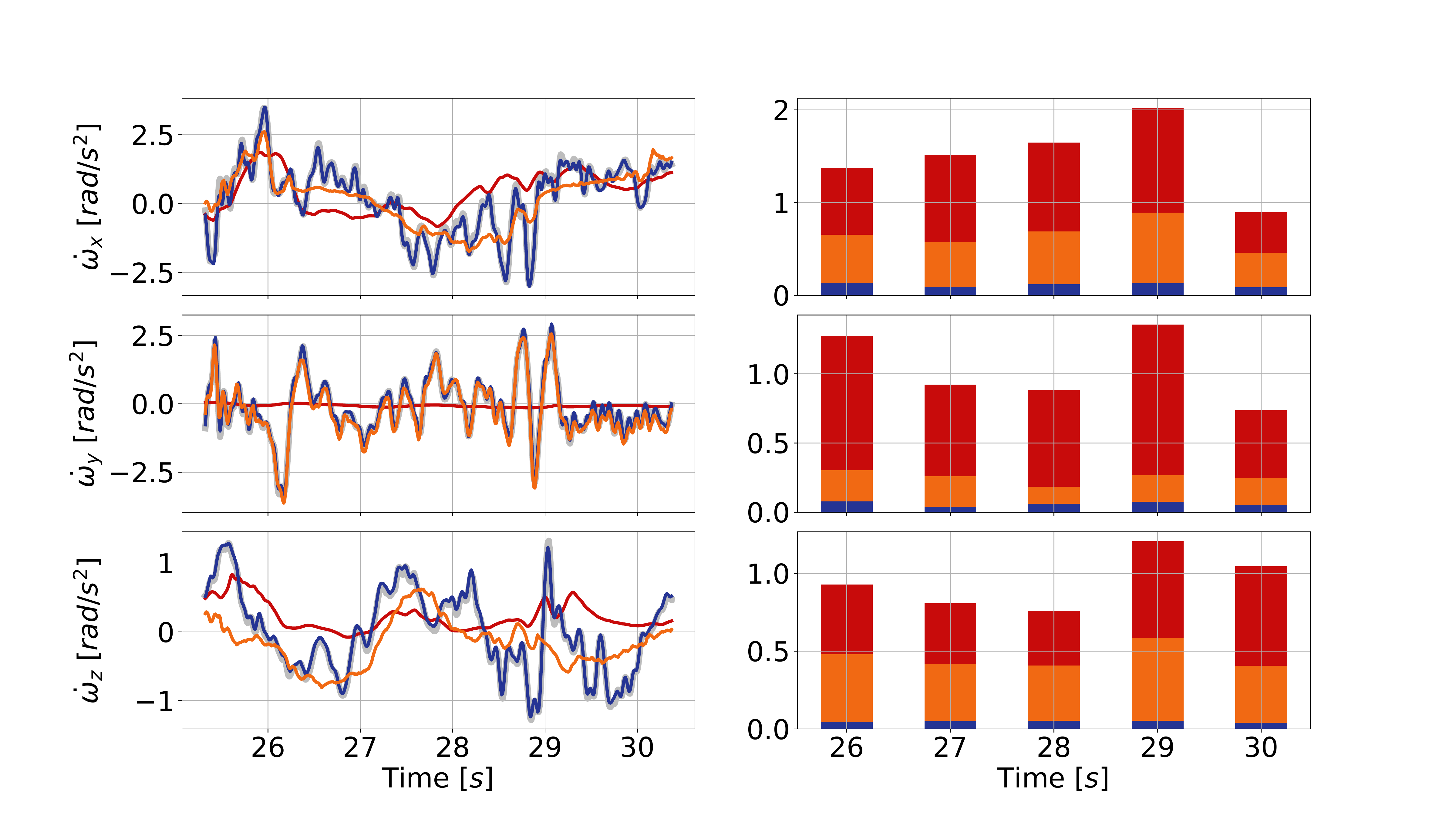}}
\end{figure*}
\begin{figure*}
\centering
\subfigure{\includegraphics[width=0.49\linewidth, trim=70 10 80 20, clip]{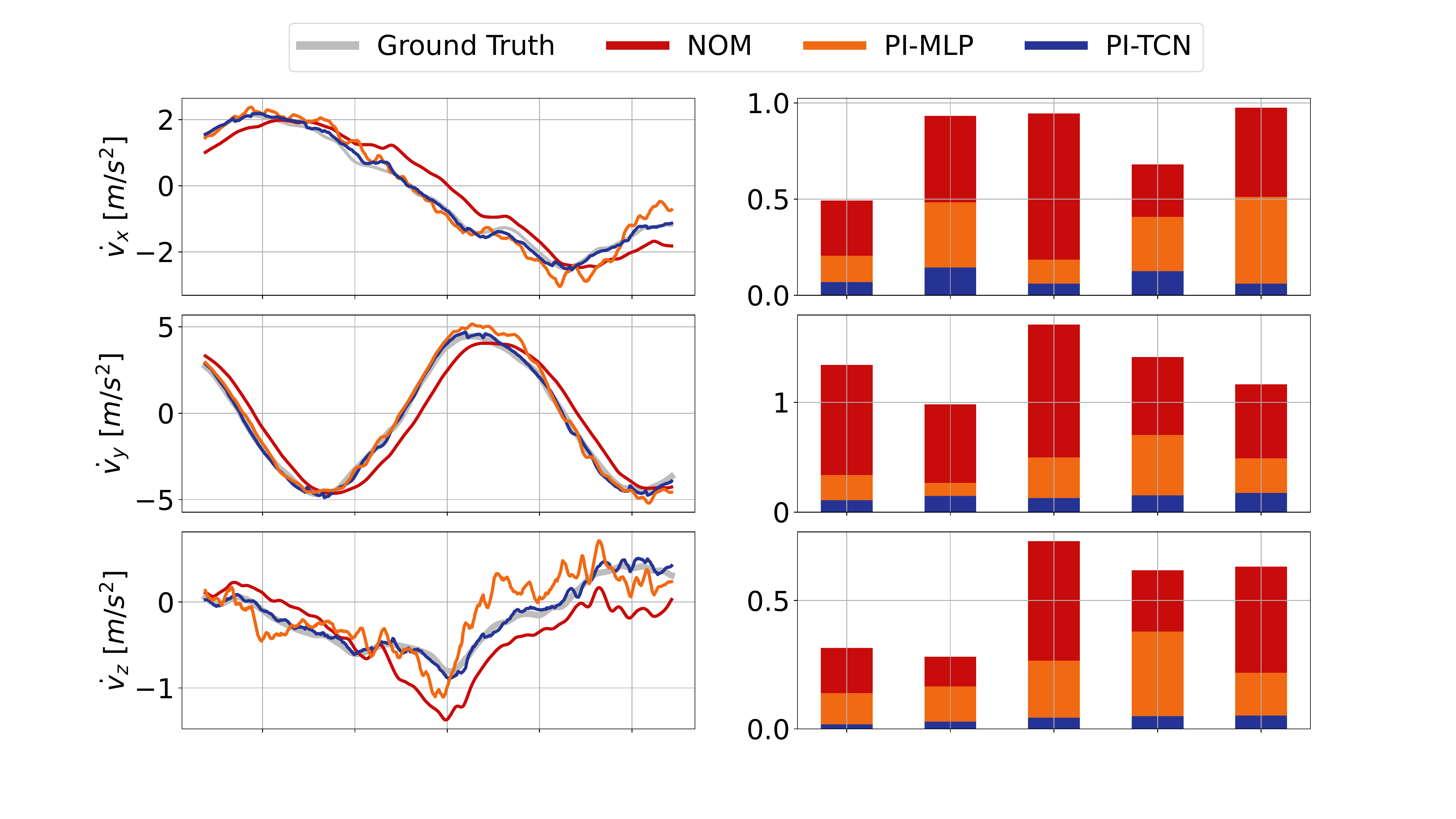}}
\subfigure{\includegraphics[width=0.49\linewidth, trim=70 10 80 20, clip]{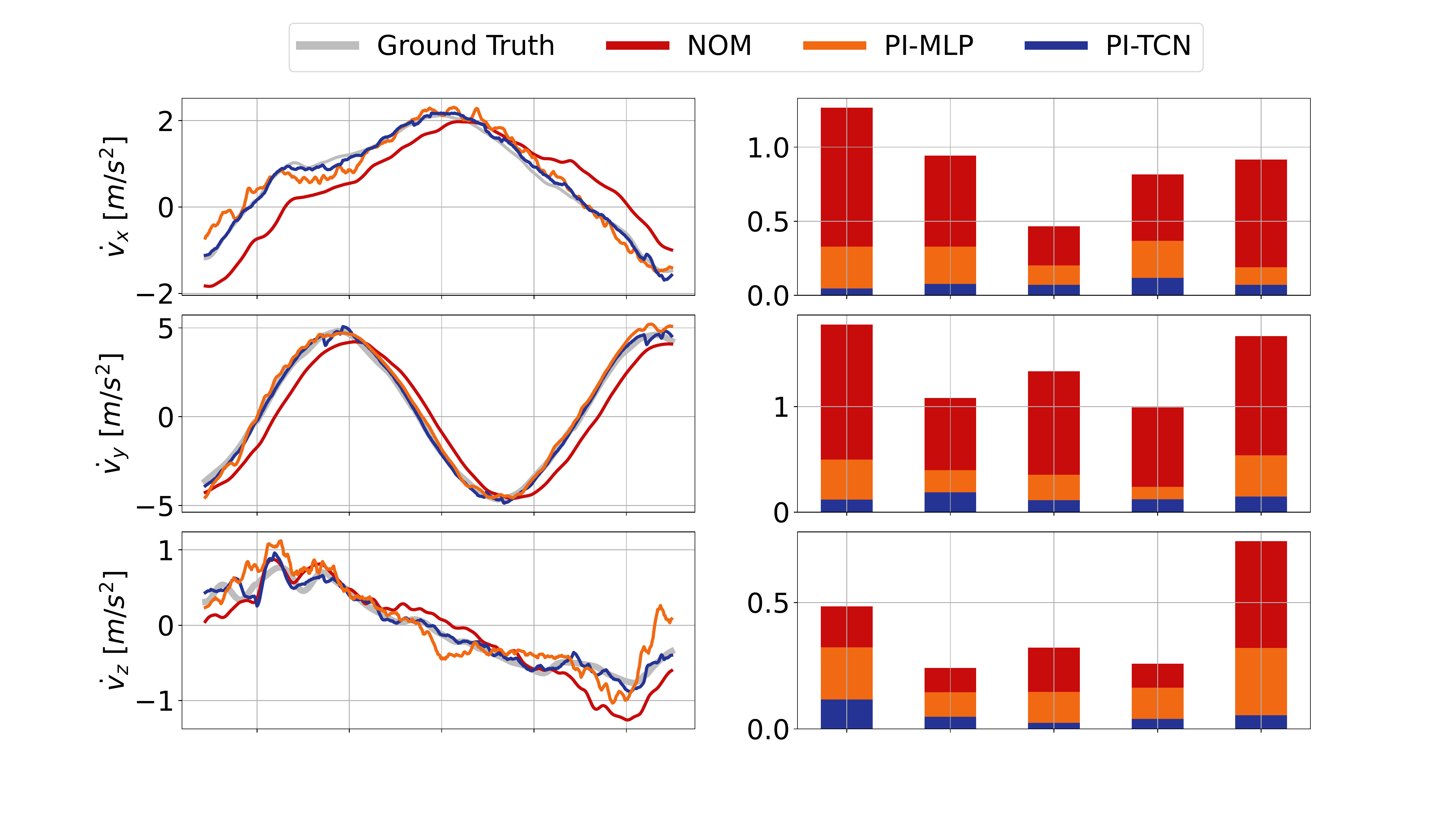}}
\vspace{-5em}
\end{figure*}
\begin{figure*}
\centering
\subfigure{\includegraphics[width=0.49\linewidth, trim=70 10 80 20, clip]{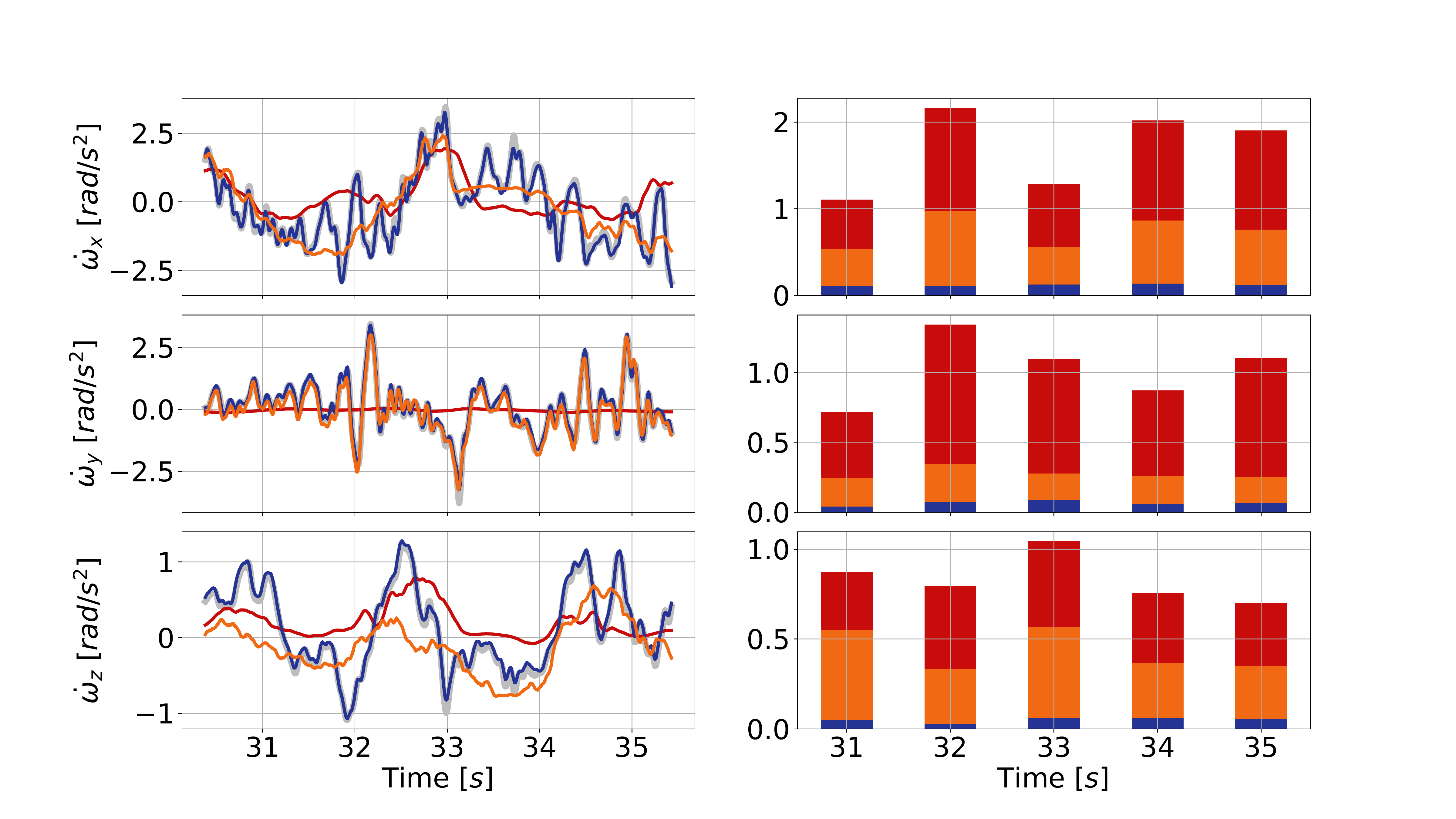}}
\subfigure{\includegraphics[width=0.49\linewidth, trim=70 10 80 20, clip]{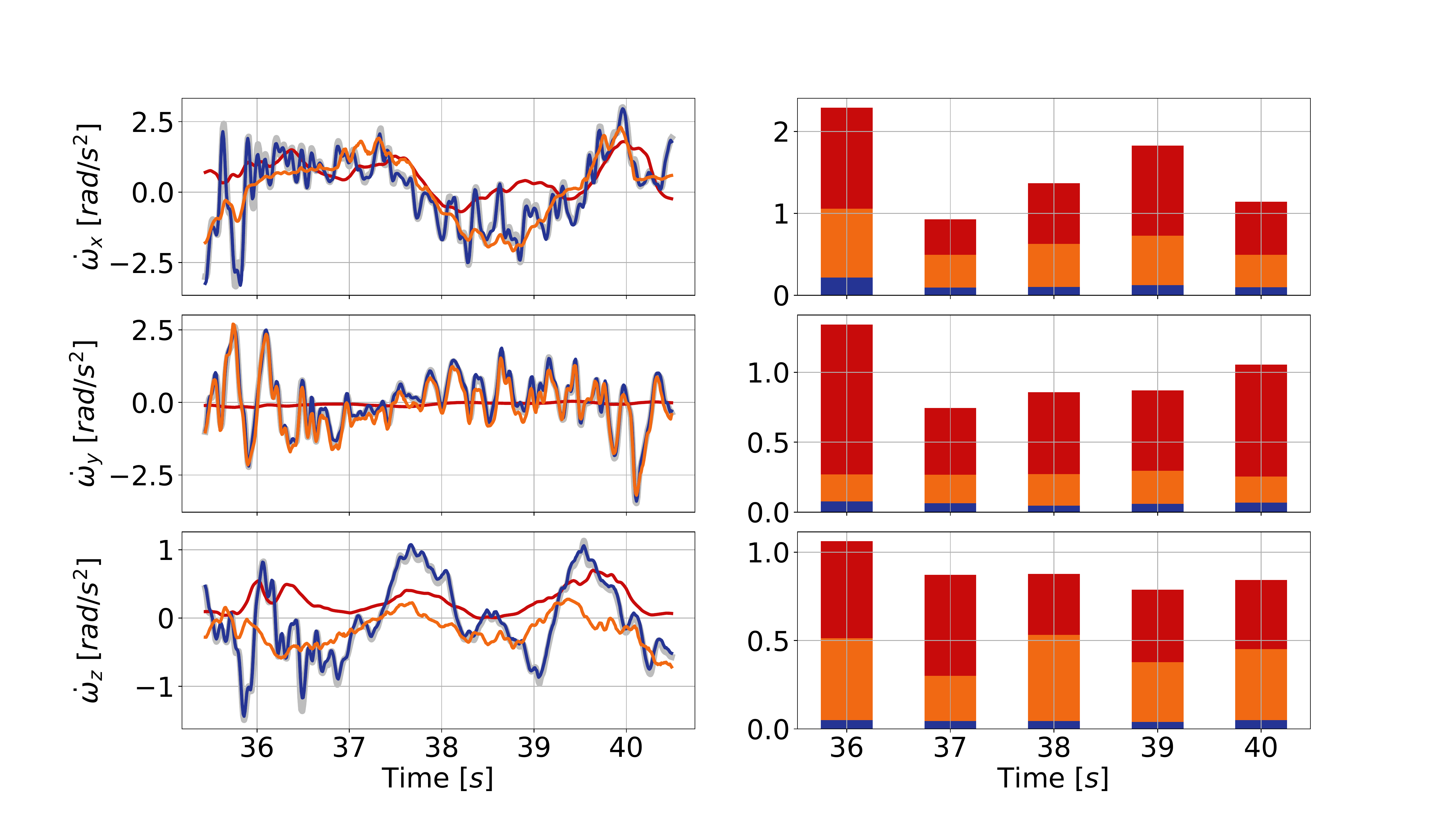}}
\caption{Predictive performance over the second half of WarpedEllipse\_1. \textbf{Left:} prediction over a slice of the trajectory. \textbf{Right:} root mean square error between prediction and ground truth (stacked vertically with no overlap).}
\label{app:pred_perf_warpedellipse_2}
\end{figure*}

\end{document}